\documentclass{article}

\usepackage[eandd, final]{neurips_2026}

\usepackage[utf8]{inputenc} 
\usepackage[T1]{fontenc}    
\usepackage{hyperref}       
\usepackage{url}            
\usepackage{booktabs}       
\usepackage{amsfonts}       
\usepackage{nicefrac}       
\usepackage{microtype}      
\usepackage{xcolor}         

\usepackage{graphicx}
\usepackage{subcaption}
\usepackage{tabularx}
\usepackage{array}
\newcolumntype{Y}{>{\centering\arraybackslash}X}

\usepackage[most]{tcolorbox}
\tcbuselibrary{breakable,skins,listings}
\usepackage{listings}
\usepackage{makecell}
\usepackage{enumitem}

\ifdefined\lessonbox
  \renewtcolorbox{lessonbox}[1][]{%
    enhanced, breakable,
    colback=white, colframe=black, boxrule=0.4pt, arc=2pt,
    boxsep=0pt,
    left=1em, right=1em,
    top=0.6ex, bottom=0.6ex,          
    before skip=6pt, after skip=10pt, 
    before upper=\parindent0pt\noindent,
    #1 
  }
\else
  \newtcolorbox{lessonbox}[1][]{%
    enhanced, breakable,
    colback=white, colframe=black, boxrule=0.4pt, arc=2pt,
    boxsep=0pt,
    left=1em, right=1em,
    top=0.6ex, bottom=0.6ex,
    before skip=6pt, after skip=10pt,
    before upper=\parindent0pt\noindent,
    #1
  }
\fi

\makeatletter
\ifdefined\variantbox
  \renewtcolorbox{variantbox}[2][]{%
    enhanced,
    breakable,
    colback=black!1,
    colframe=black!40,
    boxrule=0.3pt,
    arc=1.5pt,
    left=0.8em,right=0.8em,top=0.6ex,bottom=0.6ex,
    title={#2},
    fonttitle=\bfseries,
    boxed title style={
      colback=black!6, colframe=black!40,
      boxrule=0.3pt, arc=1.5pt,
      left=0.5em, right=0.5em, top=0.1ex, bottom=0.1ex
    },
    #1
  }
\else
  \newtcolorbox{variantbox}[2][]{%
    enhanced,
    breakable,
    colback=black!1,
    colframe=black!40,
    boxrule=0.3pt,
    arc=1.5pt,
    left=0.8em,right=0.8em,top=0.6ex,bottom=0.6ex,
    title={#2},
    fonttitle=\bfseries,
    boxed title style={
      colback=black!6, colframe=black!40,
      boxrule=0.3pt, arc=1.5pt,
      left=0.5em, right=0.5em, top=0.1ex, bottom=0.1ex
    },
    #1
  }
\fi
\makeatother

\ifdefined\gridbox
  \renewtcblisting{gridbox}{%
    enhanced, breakable, listing only, listing engine=listings,
    colback=black!3, colframe=black!60, boxrule=0.3pt, arc=1pt,
    boxsep=0pt, left=0.4em, right=0.4em, top=0.2ex, bottom=0.2ex,
    listing options={
      basicstyle=\ttfamily\scriptsize,
      columns=fullflexible, keepspaces=true,
      showstringspaces=false, aboveskip=0pt, belowskip=0pt
    }
  }
\else
  \newtcblisting{gridbox}{%
    enhanced, breakable, listing only, listing engine=listings,
    colback=black!3, colframe=black!60, boxrule=0.3pt, arc=1pt,
    boxsep=0pt, left=0.4em, right=0.4em, top=0.2ex, bottom=0.2ex,
    listing options={
      basicstyle=\ttfamily\scriptsize,
      columns=fullflexible, keepspaces=true,
      showstringspaces=false, aboveskip=0pt, belowskip=0pt
    }
  }
\fi

\usepackage{amsmath}
\usepackage{amssymb}
\usepackage{mathtools}

\usepackage{hyperref}
\usepackage{url}
\usepackage[capitalize,noabbrev]{cleveref}

\title{Distinguishing Performance From Competence in Evaluations of Humanlike Abstract Reasoning}

\author{
  Claas Beger\thanks{Correspondence to: \texttt{claasbeger@santafe.edu}} \\
  Santa Fe Institute \\
  New Mexico, USA
  \And
  Ryan Yi \\
  Santa Fe Institute \\
  New Mexico, USA
  \And
  Shuhao Fu \\
  Santa Fe Institute \\
  New Mexico, USA
  \And
  Kaleda Denton \\
  Santa Fe Institute \\
  New Mexico, USA
  \AND
  Arseny Moskvichev \\
  Advanced Micro Devices Inc. \\
  Texas, USA
  \And
  Sarah W. Tsai \\
  Sandia National Laboratories \\
  New Mexico, USA
  \And
  Sivasankaran Rajamanickam \\
  Sandia National Laboratories \\
  New Mexico, USA
  \And
  Melanie Mitchell \\
  Santa Fe Institute \\
  New Mexico, USA
}

\begin{document}

\maketitle

\begin{abstract}
  
AI reasoning models have exceeded human performance on the ARC-AGI-1 benchmark, but does that mean state-of-the-art models have the underlying competence---humanlike abstract reasoning---the benchmark was designed to test? Here we investigate the abstraction abilities of AI models using the closely related but simpler ConceptARC benchmark. Our evaluations vary input modality (textual vs.\ visual), use of external Python tools, and reasoning effort. Beyond output accuracy, we evaluate the natural-language rules that models generate to explain their solutions, enabling us to assess whether models recognize the abstractions that ConceptARC was designed to elicit. We show that the best models' rules are frequently based on less abstract, more domain-specific concepts, capturing intended abstractions considerably less often than humans. In the visual modality, AI models' output accuracy drops sharply; however, our rule-level analysis reveals that a substantial share of their rules capture intended abstractions, even as the models struggle to apply these concepts to generate correct solutions. In short, we show that using performance (accuracy) alone can substantially overestimate AI competence in textual modalities while underestimating it in visual modalities---an illustration of the risk of mistaking performance for competence.  
\end{abstract}

\section{Introduction}
The ability to quickly form abstractions and use them to reason by analogy is central to humans' remarkable capacity to generalize knowledge to novel situations \citep{carey2011concepts,hofstadter2001epilogue,lake2017building}. Many benchmarks have been designed to evaluate abstract reasoning abilities in machines \citep{foundalis2025bongard,hofstadter1995fluid,zhang2019raven}. Among the most prominent is the Abstraction and Reasoning Corpus (ARC) \citep{chollet2019measureintelligence},\footnote{In 2023 ARC was renamed ``ARC-AGI-1,'' but for simplicity we will refer to it here as ``ARC.''} which is specifically focused on reasoning with humanlike ``core-knowledge'' priors. These priors include objectness, numerosity, and basic geometric and spatial relations, as well as certain pragmatic assumptions (for example, the simplicity principle: that a simple composition of these priors, if consistent with the demonstrations, is a better description than a more complex one). 

ARC consists of a set of idealized problems that require few-shot rule-induction and analogical reasoning. As \autoref{fig:ConceptARC-Examples} shows, each puzzle (``task'') consists of a small set of \textit{demonstrations} (initial and transformed grids) and a \textit{test} grid, each ranging in size from $1\times1$ to $30\times30$, with each cell having one of 10 possible colors.  A solver is tasked with inferring a rule governing the demonstrations and applying that rule to the test input to produce a correct output grid. According to ARC's creator, Fran\c{c}ois Chollet \citeyearpar{chollet2019measureintelligence}, ARC is 
``built upon an explicit set of priors designed to be as close as possible to innate human priors. We argue that ARC can be used to measure a humanlike form of general fluid intelligence and that it enables fair general intelligence comparisons between AI systems and humans.'' 

An important distinction in cognitive science is \textit{performance} vs.\ \textit{competence} \citep{korossy1999modeling}. Performance refers to a subject's observed behavior, such as their score on a test or accuracy on a benchmark, whereas competence refers to the underlying general ability the test or benchmark is meant to evaluate.  In many tests it is possible for a subject to exhibit high performance without possessing the underlying competence, or the reverse.  Numerous papers have argued for making this distinction in AI evaluations (e.g., \citep{cowley2022framework,firestone2020performance,hu2024auxiliary,milliere2026anthropocentric}).  However, most AI studies simply report performance (accuracy) on a benchmark, assuming that a model's performance is a good proxy for its competence. This has been the case for nearly all evaluations of ARC and related benchmarks, including in the highly publicized ARC-Prize competitions (e.g., \cite{arc_prize2024}).

Here, we apply the performance vs.\ competence distinction to assess the underlying competence of three prominent reasoning models, as well as humans, in solving tasks from ConceptARC \citep{moskvichev2023conceptarc}, a benchmark in the ARC domain containing tasks organized around the same core knowledge priors that the ARC benchmark was designed to test. We find that, using standard textual encodings for tasks, the models we evaluate approach and in some cases surpass human-level performance, measured by output accuracy.  However, by using an additional probe into their behavior---namely, analyzing the concepts used in (and faithfulness of) the natural-language rules they generate to explain tasks' demonstrations---we find that  performance can \textit{overestimate} their underlying competence for the kind of humanlike abstract reasoning the benchmark is meant to evaluate. On the other hand, presenting tasks in the visual modality results in dramatic performance drops, but surprisingly we find that in this case performance can \textit{underestimate} underlying competence.  

In the following sections, we describe our experimental methodology and results, and discuss how our findings relate to three central questions: (1) How does the performance achieved by AI models on ConceptARC tasks compare to that of humans? (2) To what extent do the rules generated by AI models and by humans capture the abstractions intended by the test designers, and to what extent do they rely on unintended, less abstract patterns?  (3) How do modality (textual vs.\ visual), reasoning effort (token budget), and Python tool access affect how well models can solve these tasks via the intended abstractions?

The contributions of this paper are (1) an empirical comparison of several state-of-the-art reasoning models with humans on the ConceptARC benchmark, across textual and visual modalities, varying levels of reasoning effort, and with and without access to Python tools;  (2) new insights from this comparison into the relationship between performance and underlying competence of AI models on abstract reasoning tasks in the ARC domain; and (3) general arguments about the importance of the performance / competence distinction in AI evaluation, as well as the need for human-aligned concepts and reasoning in many areas of AI application.  Limitations of this work, discussed in detail in \autoref{sec:limitations}, include testing on a single benchmark with a subset of reasoning efforts, as well as requiring some degree of judgment in assessing the alignment between AI models' inferred rules and human concepts. 

\section{Background}
First introduced in 2019 by Chollet \citeyearpar{chollet2019measureintelligence}, the original ARC dataset consists of 1,000 tasks, of which 800 were published and the remainder held out for private evaluation. The 2024 ARC-AGI Prize competition offered a \$600,000 grand prize for any program that exceeded 85\% accuracy, measured as the fraction of correct output grids, on a 100-task private test set.\footnote{Average human performance was measured at 64\% on the somewhat easier public evaluation set \citep{legris2025comprehensive}.} The highest scoring program, with about 54\% accuracy, employed a fine-tuned LLM, extensive data augmentation, and test-time training \citep{chollet2024arc}. Following the competition, Chollet and colleagues  tested a pre-release version of OpenAI's o3 model on a different ``semi-private'' test set of 100 tasks.  This model achieved an accuracy of 76\% on its low-effort setting and 88\% on its high-effort setting \citep{chollet2025arc}.  While o3-preview was not qualified to participate in the official competition,
its performance was described as ``a genuine breakthrough, marking a qualitative shift in AI capabilities compared to the prior limitations of LLMs'' \citep{chollet2024o3}. ARC and its successor benchmarks have since become popular proxies for measuring humanlike fluid intelligence in AI systems; reporting accuracy on the original ARC dataset and successive versions has become standard for frontier AI companies when presenting the capabilities of their latest models (e.g., \citet{pichai2025gemini,anthropic2026claude}). 

\begin{figure}[t]
  \centering
 \includegraphics[width=\textwidth]{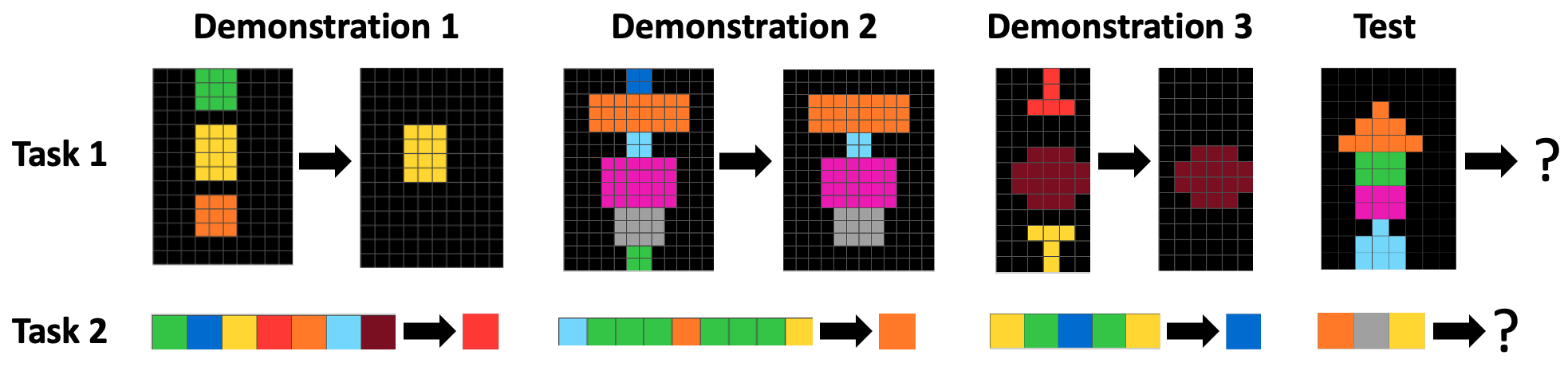}
  \caption{Two tasks from the ConceptARC benchmark.  A task consists of a small number of demonstrations of a transformation and one test grid. In our study, the solver is asked to generate a rule that describes the transformations and apply that rule to the test grid to generate an output grid.
  \label{fig:ConceptARC-Examples}}
\end{figure}

However, despite superhuman performance of reasoning models on ARC tasks, it is not clear whether these models have actually achieved the humanlike abstract reasoning and generalization competence that ARC was designed to test. Consider, for example, the task illustrated in the top row of \autoref{fig:ConceptARC-Examples} (from the ConceptARC benchmark, described below). A human solving this task could likely generalize across different instantiations of the underlying abstract concepts, namely, identifying and removing the top and bottom objects regardless of their size, shape, color, position, or number.  Yet to our knowledge, no prior studies have assessed whether AI systems such as o3 are solving these tasks by using the intended, generalizable abstractions, or whether they exploit less generalizable rules based on unintended correlations (``shortcuts'') in the demonstrations. It is well known that large neural-network models are capable of discovering spurious patterns in data and using these patterns to arrive at correct answers \citep{du2023shortcut,geirhos2020shortcut}.  Since ARC and its successors are intended as proxies for the general ability to reason abstractly with core-knowledge priors---a key ability for real-world physical understanding and for human-AI collaboration---it is essential to determine the extent to which high accuracy on these benchmarks actually corresponds to this general capacity.

Here we assess the abstractions used by three prominent reasoning models, as well as humans, in solving tasks from ConceptARC \citep{moskvichev2023conceptarc}, a benchmark in the ARC domain containing tasks organized around the same core knowledge priors that the ARC benchmark was designed to test. ConceptARC's tasks are focused on concepts such as ``inside and outside,'' ``above and below,'' or ``same vs.\ different.'' 
As described in \citet{moskvichev2023conceptarc}, ConceptARC was designed to test robust understanding of these concepts by providing tasks (designed to be simple for humans) that use each concept in various contexts and require varying degrees of generalization to solve. Because ConceptARC isolates simple abstract concepts, whereas the original ARC set may involve broader and more varied combinations of concepts, we believe ConceptARC is better suited for investigating the abstractions used by humans and machines in solving tasks. 

Nearly all ARC evaluations, including those in the ARC-AGI Prize competition, rely on the same text-based encoding of the demonstration and test grids to solve each task.  Each grid is represented as an integer matrix, with entries encoding colors indexed from 0 to 9. However, because o3 and related models are reported to possess sophisticated reasoning abilities in both textual and visual modalities \citep{OpenAIo3Vision2025}, we investigate the models' abstract reasoning abilities in both modalities. We also examine how reasoning effort (the token budget allocated for the reasoning stage) and access to external ``tools'' (here, the ability to generate and execute Python code) affect a model's ability to discover humanlike abstract rules and solve tasks.

\section{Methodology} \label{sec:Methodology}

\paragraph{Dataset and Experiments}
To create ConceptARC, Moskvichev et al.\ \citeyearpar{moskvichev2023conceptarc} chose 16 basic spatial and semantic concepts, and for each concept created 30 tasks that focused on that concept in different instantiations, with different degrees of abstraction, for a total of 480 tasks. In contrast to ARC tasks, all ConceptARC tasks were designed to be relatively easy for humans to solve: each focuses on a simple abstract concept and its application to a test grid.  

We evaluated three proprietary multimodal ``reasoning'' models on the ConceptARC tasks: OpenAI's o3, Google's Gemini 2.5 Pro, and Anthropic's Claude Sonnet 4.\footnote{For comparison, we evaluated the output accuracy of three non-reasoning multimodal models: OpenAI's GPT-4o, Meta's Llama 4 Scout, and Alibaba's Qwen 2.5 VL 72B, all of which scored much lower than the reasoning models. We also evaluated OpenAI's o4-mini for output accuracy, but chose to do further analysis on the similar-accuracy model o3. We give all of these output accuracies in \autoref{app:OutputAccuracyNonReasoningModels}.} Because the APIs for o3 and Claude Sonnet 4 restrict temperature to 1.0, we used temperature 1.0 for all three reasoning models to maintain comparability. For both textual and visual modalities, we used slight variations on the prompt used in Chollet et al.'s \citeyearpar{chollet2024o3} evaluation of o3-preview (and in all subsequent ARC-Prize evaluations of ARC-AGI-1). 
Full prompts are given in \autoref{app:textual_prompt} and \autoref{app:visual_prompt}. For each model setting, each task was evaluated in an independent API call, without prior context. 

In all experimental settings, models were asked to generate a JSON object containing the transformation rule and the corresponding output grid, represented as a matrix of integers (the same encoding used in previous ARC evaluations). This setup enables two-fold evaluation: (i) grid output accuracy and (ii) the degree to which model-generated rules capture the tasks' intended abstractions. We evaluated human-generated solutions using the same criteria, analyzing previously unpublished data obtained from the study reported in \citep{moskvichev2023conceptarc}, in which humans (participants on the Prolific Academic platform) were presented with ConceptARC tasks as images and asked to produce both the correct output grids and the rules they used to generate them.\footnote{Unpublished data was provided by A. Moskvichev.}  

Note that the natural-language transformation rules generated by each model are part of their answer to the prompt, rather than part of their chain-of-thought trace.  As we describe below, we considered these generated rules to be good proxies for the models' task-solving rationales, in that the generated rules were in almost all cases faithful to the model's generated output grid, whether correct or not.  

Due to resource constraints, we report pass@1 results for both AI models and humans.\footnote{The ARC Prize competition reported pass@2 results; that is, two independent runs on each task. If one of the runs produces a correct output grid, the task is counted as correctly solved.}  We evaluated o3 under its low- and medium-effort reasoning settings.\footnote{OpenAI does not specify the token budget allocated to these settings. Due to resource constraints, we did not test the high-effort setting.} We evaluated Gemini 2.5 Pro and Claude Sonnet 4 only in medium-effort settings, that is, we used a 16K-token reasoning budget, which roughly approximates OpenAI's medium-effort setting. Additionally, we evaluated reasoning models under two tool-access conditions: one in which Python tools were enabled and one in which they were not.

\paragraph{Evaluating Rules Generated By Humans and AI Models} For each task, humans were given the demonstrations and test grid images and were asked to generate the output grid using a custom editing tool, whereas models were asked to generate the output grid as a matrix with colors encoded as integers. Following the ARC Prize evaluation criteria, a generated output grid was considered correct only if it exactly matched the ground-truth solution in the ConceptARC corpus and was returned in the requested format. \autoref{app:alternative_grid_formats} provides a detailed analysis of cases in which models deviated from the requested format.


To investigate the extent to which AI models are using humanlike abstractions to solve tasks, we asked the models to output not only the transformed test grid but also a natural-language rule describing the transformation. (We discuss the faithfulness of these generated rules with respect to the corresponding output grids below and in \autoref{app:RuleGridAlignment}.) Moskvichev et al.\ \citeyearpar{moskvichev2023conceptarc} similarly collected such rules from their human participants, though only for correctly solved tasks.


Evaluating the correctness of natural-language rules requires human judgment. Accordingly, we manually classified both model- and human-generated rules using three possible classes: ``correct-intended'' (rules that align with the intended abstractions), ``correct-unintended'' (rules that work on the demonstrations but do not capture the intended abstractions), and ``incorrect'' (rules that do not work on the demonstrations).\footnote{In cases where the model or human participant did not provide a rule, or the rule was too unclear to confidently assess, the rule was annotated as ``non-responsive'' or ``unclear,'' respectively.} For example, in the first task shown in \autoref{fig:ConceptARC-Examples}, a sample human-generated rule is ``Use the same grid size as in the input. There will be multiple objects. Simply remove the top object and the bottom object only.''  We rated this rule as correct-intended.  Claude Sonnet 4's generated rule, from our experiment with textual inputs and medium reasoning effort, was ``Remove the topmost and bottommost colored regions, keep all colored regions in between,'' which we also rated as correct-intended.  Gemini 2.5 Pro's rule in this same setting was ``Identify all connected shapes of non-zero numbers. Sort these shapes first by their size (number of cells) in ascending order, then by their color value in ascending order as a tie-breaker. Remove the two shapes that come first in this sorted list, and keep all other shapes.''  Here, Gemini used the ordering of integer color codes as a relevant feature, even though these codes were intended to be arbitrary textual encodings.  Because Gemini identified an unintended but valid pattern in the demonstrations, we classified its rule as correct-unintended, even though the resulting output grid was incorrect.  Finally, o3's generated rule in this same setting was ``Remove all smallest objects: find each connected non-zero color component, count its cells, identify the minimum area, and delete every component having that minimum size (replace its cells with 0); leave every other cell unchanged.''  As this rule did not correctly account for all of the demonstrations, we classified it as incorrect. 

In the second task in \autoref{fig:ConceptARC-Examples}, a sample human-generated rule is ``Select the colour that is in the middle of the input. Only output that colour as a $1 \times 1$ grid.''  As before, we rated this as correct-intended. o3 with textual input, medium effort, and tool use enabled, generated the rule ``Return the middle element of the 1-D input list (index $\lfloor n / 2 \rfloor$ for odd length) as a single-value output grid.'' Also correct-intended. Claude's rule in the same setting was ``Find all values that appear with the minimum frequency in the input grid, sort them in ascending order, then return the second smallest value if multiple values exist, or the only value if just one exists.''  Although the rule correctly describes the demonstrations and happens to produce the correct output grid, its reliance on an unintended pattern rather than the relevant concept led us to classify it as correct-unintended.  In the same setting, Gemini generated a very similar correct-unintended rule, also yielding a correct output grid.  

Further examples of correct-unintended rules are given in \autoref{app:examples-correct-unintended-rules}. It is important to clarify that we do not consider such rules to be incorrect; rather, they reflect a mismatch with the intended rules---those based on humanlike abstract reasoning. 

While it is not guaranteed that a model's natural-language rule for a given task is a faithful reflection of how the task was solved, we manually analyzed the alignment between generated rules and their corresponding output grids. We found that, across all experimental settings, the output grid, whether correct or not, was faithful to the model's rule in over 90\% of the cases we examined, providing evidence that these rules serve as reasonable proxies for the model's reasoning process.  Further details are provided in \autoref{app:RuleGridAlignment}.

\section{Results}

\begin{table}[t]
  \caption{\textbf{Reasoning models:} Output-grid accuracy (pass@1) for ConceptARC for humans and models. Accuracy is shown in \%. Textual (t) and Visual (v) results are reported on separate rows, and settings with Python tools enabled are marked with +T.  Settings with ``--'' were not tested. See \autoref{sec:Methodology} for details of experiments.  
  Bold numbers correspond to the highest visual and textual scores in each column.} 
  \label{tab:arc_results_modality} 
  \centering
  \begin{tiny}
    \setlength{\tabcolsep}{4.5pt}
    \renewcommand{\arraystretch}{1.1}
    \begin{sc}
      \begin{tabular}{l@{\hspace{6pt}}c@{\hspace{8pt}}cccc}
        \toprule
        \textbf{Model} &  &
        \textbf{low} &
        \textbf{medium} &
        \textbf{low +T} &
        \textbf{medium +T} \\
        \midrule
        \textbf{o3} & \textnormal{t} &
        \textbf{68.3} & \textbf{77.1} & \textbf{67.9} & \textbf{75.6} \\
                    & \textnormal{v} &
        \textbf{6.7}  & \textbf{5.6}           & \textbf{18.1} & \textbf{29.2} \\
        \midrule
        \textbf{Claude} & \textnormal{t} &
        -- & 60.2 & -- & 55.0 \\
                        & \textnormal{v} &
        -- & 5.2  & -- & 6.9 \\
        \midrule
        \textbf{Gemini} & \textnormal{t} &
        -- & 66.0 & -- & 60.4 \\
                        & \textnormal{v} &
        -- & 4.2  & -- & 5.8 \\
        \midrule
        \midrule
        \textbf{Humans} & \textnormal{v} &
        \multicolumn{4}{c}{\hspace{-3.5em}73.0} \\
        \bottomrule
      \end{tabular}
    \end{sc}
  \end{tiny}
\end{table}

\paragraph{Output Grid Accuracy}
\autoref{tab:arc_results_modality} gives the pass@1 output-grid accuracies of the reasoning models we evaluated, in both textual and visual modalities. All three models show a substantial performance gap between the textual and visual settings. We also observe an increase in visual accuracy when Python tools are enabled, most prominently for o3 and, to a lesser extent, for Claude and Gemini. In contrast, allowing Python tools does not have a similar effect in the textual setting. For o3, increased reasoning effort is associated with increased accuracy in the textual modality, with or without tools; in the visual modality, we observe that the models primarily use the increased reasoning budget to execute more Python code, which may explain the substantial improvement in medium effort + tools. It should be noted that since these results are based on a single pass, they do not capture variation across repeated responses to the same task; we investigate that variability on a 48-task subset in the secondary analysis below.

Inspecting the failure cases of the visual setting more closely, we also find that models struggle to recognize the correct grid size from the image inputs. When Python tools are enabled, the models often use computer vision libraries to partially compensate for this difficulty. In both textual and visual modalities, the majority of incorrect outputs were well-formed grids that did not match the ground-truth solution. Particularly in the visual setting, however, a small share of outputs were not well-formed, typically because they contained uneven row lengths or non-integer tokens. \autoref{fig:Error-Type-Overview} in \autoref{app:error-type-overview} gives an error-type distribution for o3. 

Using unpublished data from Moskvichev et al.'s \citeyearpar{moskvichev2023conceptarc} study, we found that human-generated output grids achieved an overall pass@1 accuracy of 73\% on the 480 ConceptARC tasks, lower than that of o3 in the textual modality. Per-concept accuracies are reported in \autoref{app:concept_performance}.

\begin{figure}[t]
  \centering
 \includegraphics[width=\textwidth]{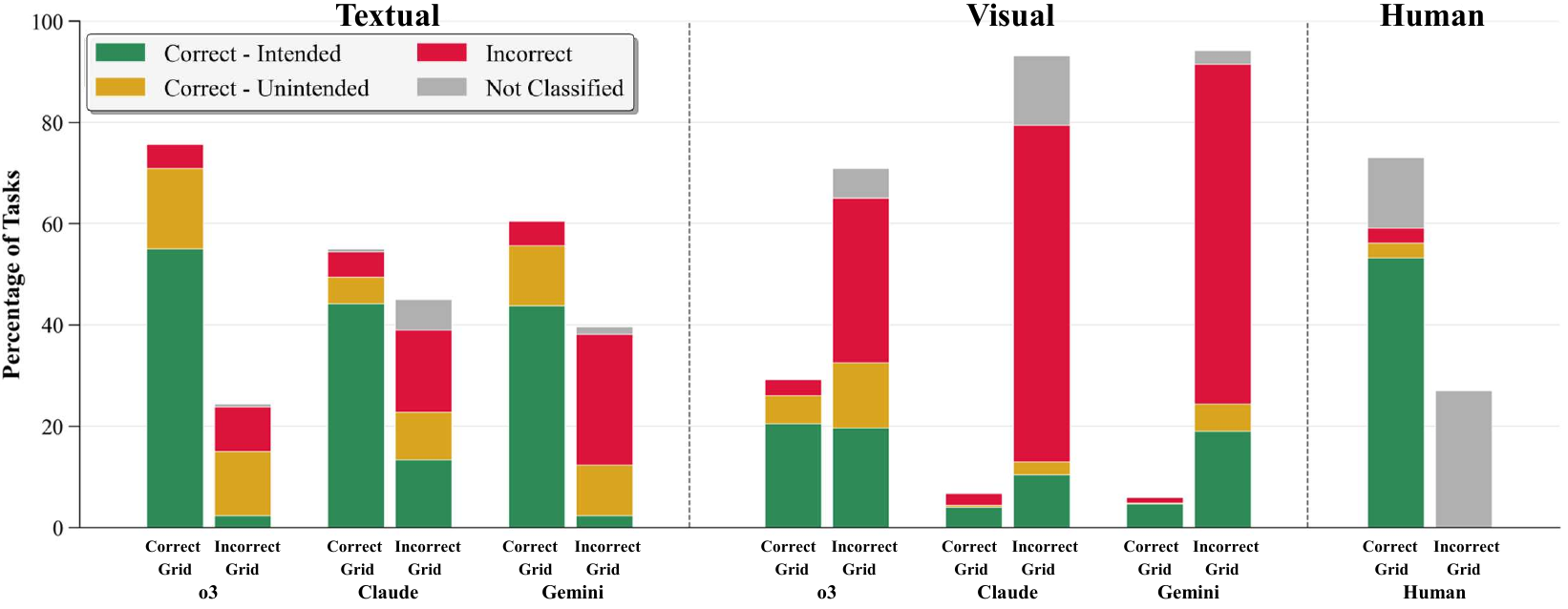}
  \caption{Results of rule evaluations.  For each model, as well as for humans, two bars are shown, representing the percentages of correct and incorrect grid outputs across the 480 ConceptARC tasks. Model evaluations are reported for both textual and visual settings.  Each bar shows the fraction of tasks for which the rule is correct-intended, correct-unintended, and incorrect.  The gray regions represent rules that could not be classified, further described in the section on Rule Evaluation. Detailed percentages for each bar are given in \autoref{app:rule_evaluation_data}. 
  \label{fig:rule_eval_plot}}
\end{figure}

\paragraph{Rule Evaluation}
Our team manually evaluated the rules generated by o3 in all settings and by Claude Sonnet 4 and Gemini 2.5 Pro in the medium-effort + tools setting, for both textual and visual modalities. We also evaluated the pass@1 rules generated by humans, using data from the study by Moskvichev et al.\ \citeyearpar{moskvichev2023conceptarc}. Classifications were made with reference to ground-truth rules provided by the ConceptARC designers. For each rule (human or machine-generated), an initial classification was assigned by one member of our team, and reviewed by a second member.  In cases where there was disagreement or uncertainty about a rule's classification, our team discussed the rule together until we came to a consensus.  

To assess the reliability of these annotations, we also conducted an inter-annotator agreement study on a subset of the corpus. Four annotators independently classified 192 rules from 48 test cases, with one rule per test case from each of o3, Claude, Gemini, and human participants. Across the six annotator pairs, mean agreement was 84.8\%, and mean Cohen's $\kappa$ was 0.70, indicating substantial agreement.

\autoref{fig:rule_eval_plot} shows the results of our rule evaluations on o3, Claude, and Gemini, all using medium-effort + tools, in both textual and visual modalities, as well as evaluations for human-generated rules.  In each setting there are two bars for each model: Correct Grid and Incorrect Grid.  The height of each bar corresponds to the percentage of the 480 tasks on which the model's output grid was correct or incorrect.  Within each bar, the green, yellow, and red sections correspond respectively to tasks for which the model's generated rule was classified as correct-intended, correct-unintended, and incorrect. Gray areas in the bars correspond to solutions for which we were unable to classify the rule, either because no rule was given by the participant or model, no rule was collected by the experimenters (this was the case for all of the human tasks with incorrect outputs), or the rule given was too unclear to confidently evaluate. Results for the textual and visual modalities are given on the left and the middle parts respectively. The rightmost part gives the output accuracy and rule evaluation results for human-generated rules.  Detailed percentages for each bar are given in \autoref{app:rule_evaluation_data}.

Notably, while o3 in the textual setting rivals humans in output grid accuracy, \textasciitilde27\% of its correct outputs are based on correct-unintended or incorrect rules, indicating reasoning based on less-abstract, often domain-specific patterns rather than intended abstract concepts. We found several types of unintended rules used by models, including rules that described complicated (and spurious) patterns in the demonstrations, rules that focused on irrelevant features such as numerical relationships among the specific numbers encoding grid colors, and rules that came close to capturing intended abstractions but included irrelevant spurious associations. In comparison, we found that \textasciitilde7\% of humans' correct outputs were based on correct-unintended or incorrect rules. 

It should be noted that our analysis of human-generated rules is limited in that roughly 19\% of the rules associated with correct outputs were unclear, non-responsive, or otherwise not classifiable, and no rules were collected for incorrect outputs, so we can compare human- and AI-generated rules only on tasks for which correct outputs were given.  Nevertheless, this subset of human- and AI-generated rules indicates that humans focus on intended concepts considerably more frequently than do AI models. This difference in focus is also clear when we analyze word frequencies in the generated rules (described below). This difference should be clarified in future human studies. Comparing AI models with one another, both Claude and Gemini have a smaller fraction of correct-unintended rules than o3, but both are lower than o3 in output accuracy.  Also notable is the percentage of incorrect output grids that are based on correct-intended rules, indicating that the models recognized the intended abstract rule, but were unable to apply it correctly to the test grid. 
In the visual setting, meanwhile, correct-intended rules were more frequent than correct output grids for all three models: 40.0\% versus 29.2\% for o3, 14.4\% versus 6.9\% for Claude, and 23.5\% versus 5.8\% for Gemini. The corresponding differences were 10.8, 7.5, and 17.7 percentage points, respectively.\footnote{The approximate paired 95\% confidence intervals were [6.2, 15.5], [4.3, 10.7], and [14.0, 21.4] percentage points for o3, Claude, and Gemini, respectively. All three gaps were significant by two-sided exact McNemar tests ($p<10^{-5}$).}
These findings illustrate the importance of going beyond accuracy in assessing the capabilities of AI models. 

Whereas \autoref{fig:rule_eval_plot} showed our rule evaluations for different models using medium reasoning effort + tools, we also analyzed the effects of varying reasoning effort and tool use for the o3 model's generated rules, in both textual and visual modalities. We found that in the visual setting, increasing reasoning effort alone does not substantially improve output accuracy or rule correctness, but enabling Python tool use does result in improvements, likely because the model is able to use computer vision libraries. In the textual setting, increasing reasoning effort has a larger positive effect on both output accuracy and rule correctness than enabling Python tool use. (For details, see \autoref{app:o3RuleAnalysis}.)

\begin{figure}[t]
  \centering
\includegraphics[width=0.65\columnwidth]{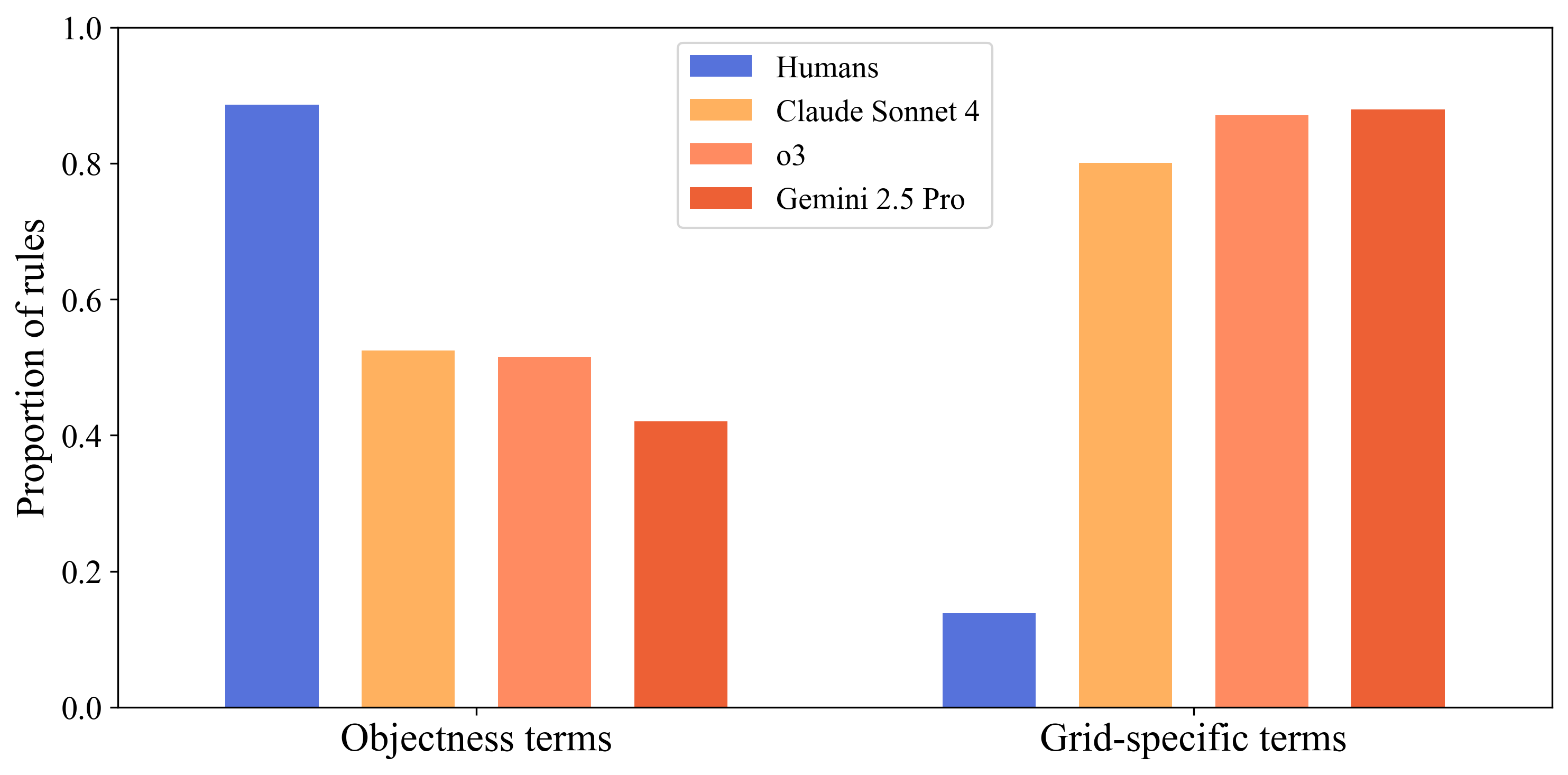}
  \caption{Proportion of rules containing at least one ``objectness" term (left) or grid-specific term (right), for humans and AI models. 
  \textbf{Objectness terms:} object, shape, square, rectangle, rectangular, line, triangle, triangular, pyramid, diamond, cross, polygon.
  \textbf{Grid-Specific terms:} cell, pixel, row, column, value, number, neighbor, neighborhood, index, subgrid, coordinate. The number of rules analyzed for humans, Claude, o3, and Gemini was 2456, 261, 363, and 290, respectively.  
  }
  \label{fig:per_llm_proportions_with_tools}
\end{figure} 

\paragraph{Comparing Concepts Used In Human and AI Model Rules \label{sec:ComparingConcepts}}
A fundamental aspect of ``core knowledge'' is the notion of \textit{objectness}---the compositional parsing of situations into objects, object attributes, and relations. Many, if not most, tasks in the original ARC dataset and ConceptARC were intended to involve abstractions related to objectness.  However, we observed that the rules generated by AI models often focused on individual grid cells, rows or columns, and other domain-specific features, rather than on more generalizable abstract concepts. To quantify these differences, we manually inspected the rules generated by humans and models and identified two sets of words: one set associated with objectness, and another associated with more surface-level, grid-specific concepts. We then used a Python script to normalize word forms and, for each set of terms, count the number of rules containing at least one term from that set. A rule could be counted in both categories. Because human rules were available only for tasks with correct output grids, we likewise restricted the model rules to cases with correct grids. We also excluded unclear and non-responsive rules, as well as all rules from visual settings, as there were too few correct outputs to support a meaningful comparison. \autoref{fig:per_llm_proportions_with_tools} gives the proportion of human- and LLM-generated rules that contain at least one term from each set given in the caption, showing that human rules were substantially more likely to contain objectness-related terms and less likely to contain grid-specific ones, while the opposite is true for AI models. These results are consistent with our broader findings that AI models are substantially less likely than humans to rely on intended abstractions when solving ConceptARC tasks. Details of this analysis and similar analyses of terms used in rules for each of our rule classifications (Correct-Intended, Correct-Unintended, and Incorrect) and for each concept-group are given in \autoref{app:ConceptUseComparison}. We consider this a preliminary analysis because selecting the objectness and grid-specific terms involved some amount of judgment and manual inspection; future work should investigate more principled and automated methods.

\paragraph{Repeated-Run Analysis}
Finally, because the main results above are based on a single pass, we also assessed how much model responses varied across independent runs. We ran each model (textual, medium-effort, without tools) three times on a fixed subset of 48 test cases. Aggregate output accuracy and rule-category rates were stable across the three passes, although individual responses exhibited some variation. The same rule classification appeared in all three passes for 54.2--56.4\% of eligible test cases across models, while average pairwise rule-label consistency ranged from 66.0\% to 70.9\%. Our results (and the specific cases discussed above) should therefore be interpreted as characterizing general patterns observed across model responses, rather than describing deterministic strategies for a given model on a particular task. Detailed methods, rates by individual model, 95\% bootstrap confidence intervals, and stability results are reported in \autoref{app:repeated-runs}. 

\section{Discussion}
In reference to the questions we listed at the beginning of this paper: (1) How does the accuracy obtained by AI models compare with that of humans? \autoref{tab:arc_results_modality} shows that for textual inputs and medium reasoning effort, o3 surpasses human accuracy on ConceptARC tasks, with Claude and Gemini obtaining lower accuracy. This aligns with results reported in \citet{chollet2025arc} and \citet{arc_agi_leaderboard}.
However, the models' performance in the visual modality is considerably worse. (2) To what extent do the rules generated by AI models and by humans capture the abstractions intended by the test designers?  For textual inputs, with medium reasoning effort and Python tools enabled,  \textasciitilde27\% of o3 and Gemini's rules associated with correct outputs were correct-unintended or incorrect; for Claude this percentage was  \textasciitilde19\% and for humans \textasciitilde7\%.
Moreover, \autoref{fig:per_llm_proportions_with_tools} and \autoref{fig:per_correctness_proportions_with_tools} show that humans' rules were considerably more likely to contain objectness-related terms (even for correct-unintended and incorrect rules), while AI models' rules were considerably more likely to contain domain-specific grid-related terms.  These results demonstrate that AI models are substantially more likely than humans to miss intended abstractions and instead solve tasks using less humanlike strategies, examples of which are detailed in \autoref{app:examples-correct-unintended-rules}.  

One might argue that if humans were given a textual rather than visual representation of these tasks, they might also use some of the same shortcuts that we saw in AI models' rules. This has not been tested, but our point here is different: in past work on ARC, high performance using this same textual encoding has prompted claims that AI models are reasoning with the human-like abstractions that ARC is meant to test. It is this assumption---that high performance implies high competence---that we are arguing against here. With respect to the original ARC-AGI-1 benchmark and its successor ARC-AGI-2 \cite{chollet2025arc}, our team performed preliminary experiments that indicate overfitting of AI models to the standard numerical encoding of tasks for these benchmarks as well; our results are reported in \autoref{app:ARC_AGI_Overfitting}.

(3) Regarding the effects of textual versus visual modalities,
\autoref{tab:arc_results_modality} and \autoref{fig:rule_eval_plot} show that while both output-grid and rule correctness drop dramatically in the visual mode, all three models are better at forming correct-intended rules than at generating correct output grids. In addition, \autoref{tab:arc_results_modality} shows that increased reasoning effort is more helpful for textual inputs, whereas Python tools are more helpful for visual inputs.
These results point to possible directions for strengthening visual reasoning models, especially in more abstract domains.  


Our results highlight the importance of distinguishing performance from competence, namely, assessing the extent to which a system's performance relies on generalizable mechanisms rather than less abstract, more domain-specific patterns \citep{frank2023baby,ivanova2025evaluate,rane2024principles}.  While in some cases it is desirable for AI systems to reason in non-humanlike ways (e.g., in discovering novel patterns in scientific data, such as protein sequences \citep{jumper2021highly}), developing AI models that grasp the abstractions understood by humans will be essential for these systems to generalize and explain their reasoning in ways understandable to humans---key abilities for successful human-AI interaction.

\section{Related Work}
\label{sec:RelatedWork}

Prior computational methods for solving ARC tasks have included program synthesis with and without pretrained LLMs \citep{banburski2020dreaming,bober2024neural,windARC2020a}, fine-tuning LLMs with augmented data and test-time training \citep{chollet2024arc}, and using fine-tuned large reasoning models \citep{chollet2024o3}. Almost all of these studies rely on textual task representations (\citet{hu2025arc} is an exception), and all focus exclusively on output-grid accuracy; to our knowledge, none analyze the rules generated by models, as was done in our study.  \citet{legris2025comprehensive} collected rules generated by humans and trained a Naive Bayes classifier to predict which task in their study was described by a particular rule, finding that humans indeed used expected core knowledge priors (i.e., concepts related to objectness, basic geometry and topology, and numerosity) in their rules.

Among other benchmarks that have been used to evaluate abstract and visual reasoning abilities, those closest to ARC and ConceptARC include Bongard problems \citep{Bongard1970}, letter-string analogies \citep{hofstadter1985metamagical}, Raven's progressive matrices (RPMs) \citep{raven1938}, and compositional visual reasoning (CVR) \citep{zerroug2022benchmark}. Bongard problems are similar to ConceptARC tasks in that each problem tests understanding of a single core spatial or semantic concept, such as ``large vs. small'' or ``inside vs. outside.''  Bongard problems are inherently visual, and several studies have evaluated multimodal models on subsets or variations of the original problems, finding that, consistent with our results on ConceptARC, the poor performance of VLMs on these problems appears to arise primarily from difficulties with vision rather than with reasoning \citep{malkinski2024reasoning,pawlonka2025bongard,wust2024bongard}. Letter-string analogies were first proposed by \citet{hofstadter1985metamagical} as an idealized domain for analogy-making.  \citet{webb2023emergent} found that GPT-3 reached human level accuracy on letter-string analogies, but other studies (testing both GPT-3 and GPT-4) found that LLMs were not robust to problem variations that did not affect humans' performance \citep{Hodel2023,lewis2025evaluating}.   Raven's progressive matrices (RPMs) have long been used as tests of human fluid intelligence. RAVEN, a dataset consisting of programmatically generated, simplified RPMs \citep{zhang2019raven}, has been used to evaluate visual reasoning in VLMs (e.g., \citep{zhang2024far,zhu2025data}), which seem to struggle more with perceptual understanding than with reasoning. The CVR benchmark \citep{zerroug2022benchmark} tests compositional visual reasoning abilities based on concepts similar to those used in ConceptARC. CVR has been used to evaluate convolutional neural networks as well as vision transformers, none of which approach human-level performance.


\section{Conclusions}

In this work we make the following contributions: (1) We demonstrate the effects of task representation (textual or visual), reasoning effort, and Python tool use on the ConceptARC benchmark, finding that in textual modalities with medium reasoning effort, the best AI models approach or surpass humans in output accuracy. (2) We further evaluate the rules that AI models generated to describe their solutions, finding that, compared with humans, their rules more frequently rely on less generalizable features or patterns. These results suggest that relying on performance alone to evaluate abstract reasoning capabilities, as was done in the ARC-Prize challenge as well as many other evaluations of AI abstract reasoning, can substantially overestimate general competence.
(3) We show that the multimodal reasoning models we tested perform dramatically worse on visual versus textual inputs.  However, in visual settings, these models were better at generating correct-intended rules than they were at applying them, again illustrating the limitations of performance alone as a proxy for general competence, and pointing to potential directions for improving AI visual reasoning. In terms of broader impacts, improving the abilities of AI models to recognize and use humanlike abstract concepts is a crucial step toward making AI systems more generalizable and trustworthy in human-centered domains, as well as enabling them to effectively work with and communicate their reasoning processes to humans. To facilitate follow-up research, we provide all of the data reported in this paper on Hugging Face,\footnote{\url{https://huggingface.co/datasets/ClaasBeger/ConceptARC_Rule_Annotations}} along with a web interface to view it interactively.\footnote{\url{https://claasbeger.github.io/performance-competence-gap/}}

\section{Limitations} \label{sec:limitations}
Due to resource limitations, we did not experiment with higher-effort reasoning settings, and we report pass@1 accuracies for both humans and machines rather than the pass@2 or pass@3 accuracies reported in other ARC evaluations. In addition, our manual (though team-based) classification of human- and machine-generated rules necessarily involved some interpretive judgment, and would be difficult to scale to larger datasets.  We are not aware of any objective or algorithmic means to usefully classify these natural-language rules into our classification categories; an important future direction is to develop AI systems to be reliable judges for such a task.   

Additionally, the prompts used in our experiments were intentionally kept as close as possible to those used in the ARC Prize evaluation of o3 \citep{chollet2024o3}; alternative prompts may elicit stronger performance from these systems. The data we obtained for human-generated rules were incomplete: no rules were collected for incorrect outputs, and even among correct outputs, some human-generated rules were not classifiable for reasons described earlier in the paper. Another limitation is that our study involves only the ConceptARC dataset; it is possible that tasks in the original ARC test sets are more resistant to the kinds of rule ``shortcuts'' seen in our results. However, to our knowledge, there has been no prior research on this topic beyond the preliminary study we report in \autoref{app:ARC_AGI_Overfitting}. Lastly, we did not evaluate newer models that have been released since the o3, Claude, and Gemini evaluations we report here; however, the arguments we make here on distinguishing performance from competence remain relevant to any such evaluation.

\newpage


\begin{ack}
This research was supported by the BANYAN Institute, funded by Sandia National Laboratories’ Laboratory Directed Research and Development program, and by the grant ``Building Diverse Intelligences through Compositionality and Mechanism Design'' from the Templeton World Charity Foundation, https://doi.org/10.54224/33804 and https://doi.org/10.54224/20650.  

Sandia National Laboratories is a multimission laboratory managed and operated by National Technology and Engineering Solutions of Sandia, LLC., a wholly owned subsidiary of Honeywell International, Inc., for the U.S Department of Energy’s National Nuclear Security Administration under contract DE-NA-0003525.

Arseny Moskvichev is an employee of Advanced Micro Devices Inc. The authors declare no other competing interests.
\end{ack}

\bibliography{references}
\bibliographystyle{plainnat}

\newpage


\appendix
\onecolumn

\section{Textual Prompt \label{app:textual_prompt}}
\begin{lessonbox}[top=0.6ex]
Find the common rule that maps an input grid to an output grid, given the examples below.

\medskip
\textbf{Example 1}

\emph{Input:}
\begin{gridbox}
0 0 0 0 0 0 0 0 0 0 0 0
0 0 0 4 4 4 4 0 0 0 0 0
0 0 0 4 4 4 4 0 0 0 0 0
0 0 0 4 4 4 4 0 0 0 0 0
0 0 0 0 0 0 0 0 0 0 0 0
0 0 0 0 0 0 0 0 0 0 0 0
2 2 2 2 2 2 2 2 2 2 2 2
0 0 0 0 0 0 0 0 0 0 0 0
0 0 0 4 4 4 4 0 0 4 4 4
0 0 0 4 4 4 4 0 0 4 4 4
0 0 0 4 4 4 4 0 0 4 4 4
\end{gridbox}

\emph{Output:}
\begin{gridbox}
0 0 0 0 0 0 0 0 0 0 0 0
0 0 0 4 4 4 4 0 0 0 0 0
0 0 0 4 4 4 4 0 0 0 0 0
0 0 0 4 4 4 4 0 0 0 0 0
0 0 0 0 0 0 0 0 0 0 0 0
0 0 0 0 0 0 0 0 0 0 0 0
2 2 2 2 2 2 2 2 2 2 2 2
0 0 0 0 0 0 0 0 0 0 0 0
0 0 0 0 0 0 0 0 0 0 0 0
0 0 0 0 0 0 0 0 0 0 0 0
0 0 0 0 0 0 0 0 0 0 0 0
\end{gridbox}

\medskip
\textbf{Example 2}

\emph{\textit{Abbreviated}}

\medskip
\textbf{Example 3}

\emph{\textit{Abbreviated}}

\begin{tcolorbox}[raster columns=2, raster equal height,
  colback=white, colframe=white, boxrule=0pt, boxsep=0pt,
  left=0pt, right=0pt, top=0pt, bottom=0pt]
  \begin{variantbox}{No Tools Variant}
Below is a test input grid. Predict the corresponding output grid by applying the rule you found. Do not generate any Python code or use any external tools to solve this task.
  \end{variantbox}
  \begin{variantbox}{Tools Variant}
Below is a test input grid. Predict the corresponding output grid by applying the rule you found. Use python if needed.
  \end{variantbox}
\end{tcolorbox}

\medskip
\emph{Test Input:}
\begin{gridbox}
0 6 6 0 0 6 6 6 0 6
0 6 6 0 0 6 6 6 0 6
1 1 1 0 0 6 6 6 0 0
4 4 4 4 4 4 4 4 4 4
0 0 0 0 0 0 0 0 0 0
0 0 0 0 0 0 0 0 0 0
0 6 6 6 0 0 6 6 6 0
0 6 6 6 0 0 6 6 6 0
0 6 6 6 0 0 0 0 0 0
0 0 0 0 0 0 0 0 0 0
0 0 0 0 0 0 4 4 4 4
0 6 6 6 6 0 0 0 0 0
0 6 6 6 6 0 0 0 0 0
0 6 6 6 6 0 0 0 0 0
\end{gridbox}

\medskip
\textbf{Return only this minified JSON (no markdown, no extra keys):}
\begin{gridbox}
{"rule":"<Transformation rule>","grid":"<final grid>"}
\end{gridbox}
\end{lessonbox}
\section{Visual Prompt \label{app:visual_prompt}} 
\begin{lessonbox}[top=0.6ex, before upper=\parindent0pt\noindent]
The left side of the first image shows 3 grids, where each grid square is colored with one of 10 possible colors: black, blue, red, green, yellow, gray, magenta, orange, cyan or brown.
The right side of the first image also contains 3 grids, each of which is a transformed version of the corresponding grid on the left. There is a single rule that describes these 3 transformations.

\begin{tcolorbox}[raster columns=2, raster equal height,
  raster before skip=0pt, raster after skip=0pt,
  colback=white, colframe=white, boxrule=0pt, boxsep=0pt,
  left=0pt, right=0pt, top=0pt, bottom=0pt]
  \begin{variantbox}{No Tools Variant}
Determine the rule, and then apply that rule to the grid in the second image. Do not generate any Python code or use any external tools to solve this task.
  \end{variantbox}
  \begin{variantbox}{Tools Variant}
Determine the rule, and then apply that rule to the grid in the second image. Use python if needed.
  \end{variantbox}
\end{tcolorbox}

You may describe the final grid through natural language using the indices of the different colors, for example:
\begin{gridbox}
0 0 0 0 0 0
1 1 1 1 1 1
0 0 0 0 0 0
4 4 4 4 4 4
0 0 0 0 0 0
\end{gridbox}
which would be a 6x5 grid with a horizontal blue line in row 2 and a horizontal yellow line in row 4. The rest of the grid is black.

\begin{tcolorbox}[raster columns=2, raster equal height,
  raster before skip=0pt, raster after skip=0pt,
  colback=white, colframe=white, boxrule=0pt, boxsep=0pt,
  left=0pt, right=0pt, top=0pt, bottom=0pt]
  \begin{tcolorbox}[colback=black!1, colframe=black!30, boxrule=0.3pt, arc=1.5pt,
    left=0pt, right=0pt, top=0pt, bottom=0pt, boxsep=0pt, valign=center]
    \centering
    \includegraphics[width=\linewidth,height=0.20\textheight,keepaspectratio]{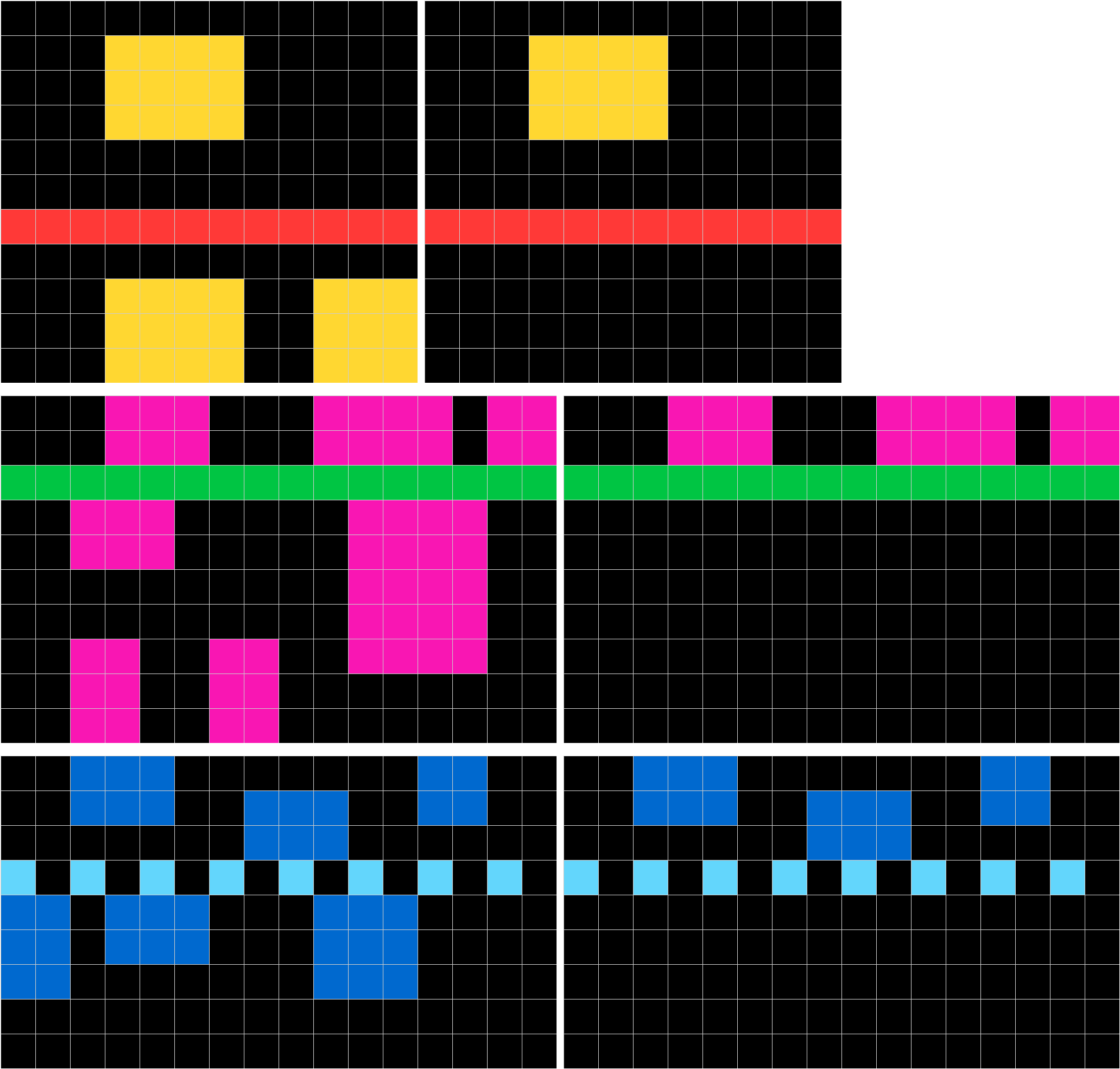}\\[-0.25ex]
    \emph{Image 1: Training examples}
  \end{tcolorbox}
  \begin{tcolorbox}[colback=black!1, colframe=black!30, boxrule=0.3pt, arc=1.5pt,
    left=0pt, right=0pt, top=0pt, bottom=0pt, boxsep=0pt, valign=center]
    \centering
    \includegraphics[width=\linewidth,height=0.20\textheight,keepaspectratio]{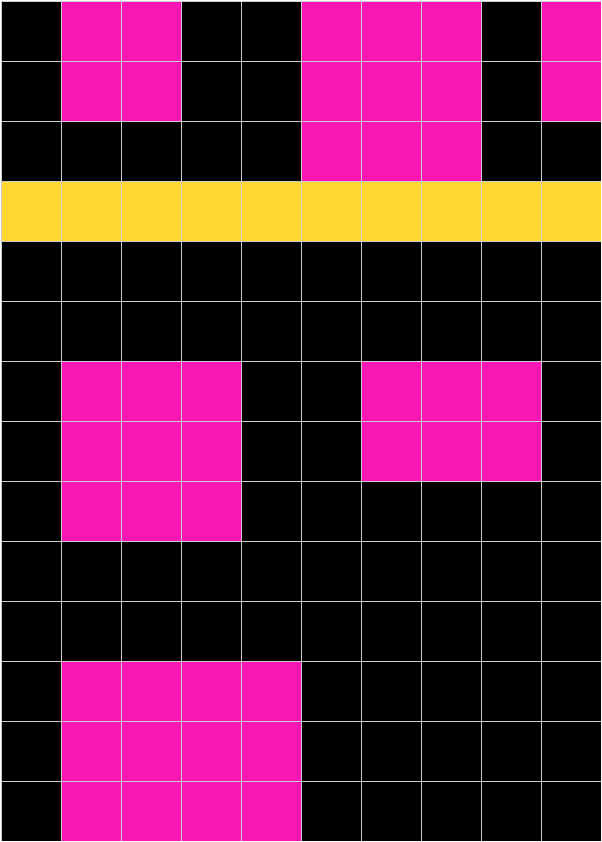}\\[-0.25ex]
    \emph{Image 2: Test grid}
  \end{tcolorbox}
\end{tcolorbox}

\medskip
---
Return \textbf{only} this minified JSON object:
{``rule":``⟨Transformation rule⟩",``grid":``⟨final grid⟩"}
No markdown, no extra keys, no code fences.
---
\end{lessonbox}
\clearpage
\section{Prompt for Non-Reasoning Models \label{app:nonreasoning}}
The prompt we used for non-reasoning models was minimally modified to require an additional field containing a reasoning trace in the final JSON object. Otherwise, the prompt was consistent with those used for reasoning models, including variations for visual settings and settings with tools enabled.

\section{Output Accuracy for o4-mini and Non-Reasoning Models} \label{app:OutputAccuracyNonReasoningModels}
o4-mini was run with medium reasoning effort, and, as with the other reasoning models, temperature 1.0. Non-reasoning models were run with temperature 0 (our tests with higher temperature yielded worse performance). As shown in \autoref{tab:arc_results_non_reasoning}, the accuracies of the non-reasoning models were dramatically lower than those of the reasoning models.  o4-mini's accuracy was similar to that of o3. For GPT-4o, in almost all cases in both modalities, the model generated an output grid that was incorrect.  For Llama 4 Scout and Qwen 2.5 VL 72B, the same was true in the textual modality; however, for Qwen, in almost all cases in the visual modality, the model was not able to generate an answer at all and did not return the requested JSON format.  This was true to a lesser extent for Llama 4 Scout.  Determining why these models had difficulty generating answers in a valid format is a topic for future research.  

\begin{table}[h]
  \caption{\textbf{o4-mini (medium effort) and non-reasoning models:} Output-grid accuracy (pass@1) on Concept-ARC across models and experimental settings. Accuracy is shown in \%. Each cell shows accuracy in the \textit{textual / visual} modality. Temperature is set to 1.0 for o4-mini and 0.0 for non-reasoning models. The Llama and Qwen interfaces did not provide options for Python tool use.}
\label{tab:arc_results_non_reasoning}
  \centering
  \small
  \setlength{\tabcolsep}{6pt}
  \renewcommand{\arraystretch}{1.25}
  \begin{tabularx}{\textwidth}{lYY}
    \toprule
    \textbf{Non-Reasoning model} &
    \thead{\textbf{No Python Tools}} &
    \thead{\textbf{With Python Tools}} \\[2pt]
    & Textual / Visual & Textual / Visual \\[2pt]
    \midrule
        \textbf{o4-mini} 
        & \textnormal{70.8} / \textnormal{8.1} & \textnormal{77.7} / \textnormal{25.0} \\        
    \midrule
    \textbf{GPT-4o} 
      & \textnormal{14.6} / 0.0 & \textnormal{8.3} / \textnormal{0.2} \\
    \midrule
    \textbf{Llama 4 Scout} 
      & 6.7 / 0.0 & - \\
    \midrule
    \textbf{Qwen 2.5 VL 72B} 
      & 9.2 / 0.0 & - \\
    \bottomrule
  \end{tabularx}
\end{table}

\section{Output Grid Accuracies Reassessed For Incorrect Grid Formats \label{app:alternative_grid_formats}}

To compute the accuracies reported in \autoref{tab:arc_results_modality} and \autoref{tab:arc_results_non_reasoning}, we followed the ARC-Prize evaluation method \cite{arc_evaluation}: we counted an output grid as correct only if it perfectly matched the ground-truth output grid and was in the format requested in the prompt (see \autoref{app:textual_prompt} and \autoref{app:visual_prompt}). However, upon exhaustive examination of the output grids generated by different models, we found that, in some cases, models generated these answer grids in different formats than that requested in the prompt; these answers were assessed as incorrect.  The incorrect output grid formats included surrounding grid rows with brackets, using commas or slashes as row separators, and several other variations.

We re-assessed each case of such formatting to see if the \textit{intended} grid was actually correct.  \autoref{tab:alternate_grid_accuracies} gives, for each model and experimental setting, the original output-grid accuracy from \autoref{tab:arc_results_modality}  or \autoref{tab:arc_results_non_reasoning} and the revised output-grid accuracy when incorrect formats are allowed. \autoref{tab:alternate_grid_accuracies} shows that accepting alternate grid formats leads to minor increases in accuracy in most cases, with a few exceptions in which the accuracy rose by more than 5\%: o4-mini low-effort, o4-mini low-effort + tools, and Claude Sonnet 4 medium-effort, which had the largest increase: 60.2\% to 72.5\%.


In summary, while models sometimes generate their answer grid in a different format than what we requested, whether we accept these formats as valid answers and assess their correctness does not have a large effect on our overall results.

In a smaller number of cases, all in the visual setting, models would generate a natural-language description of the output grid rather than the grid itself.  We did not consider these to be in a valid answer format and counted such outputs as incorrect.

\begin{table}[htbp]
  \caption{\textbf{Output grid accuracies with incorrect grid formats included.}  For each model and setting, we give \textit{original accuracy} / \textit{re-assessed accuracy}.  Original accuracies are from \autoref{tab:arc_results_non_reasoning} and \autoref{tab:arc_results_modality}.}
  \label{tab:alternate_grid_accuracies}
  \centering
  \small
  \setlength{\tabcolsep}{6pt}
  \renewcommand{\arraystretch}{1.25}
  \begin{tabularx}{\textwidth}{llYY}
    \toprule
    \textbf{Model} & \textbf{Setting} &
    \textbf{Textual} &
    \textbf{Visual} \\
    & & \thead{Original / Re-assessed} & \thead{Original / Re-assessed} \\
    \midrule
    \textbf{o3}
      & \textbf{low effort} & 68.3 / 69.4 & 6.5 / 6.5\\
    \textbf{o3}
      & \textbf{medium effort} & 77.1 / 77.1 & 5.6 / 5.6\\
    \textbf{o3}
      & \textbf{low effort + tools} & 67.9 / 68.1 & 18.2 / 18.2\\
    \textbf{o3}
      & \textbf{medium effort + tools} & 75.6 / 75.6 & 29.2 / 29.2\\
    \midrule
    \textbf{o4-mini}
      & \textbf{low effort} & 52.1 / 59.6 & 3.8 / 3.8\\
    \textbf{o4-mini}
      & \textbf{medium effort} & 70.8 / 73.8 & 8.1 / 8.1\\
    \textbf{o4-mini}
      & \textbf{low effort + tools} & 57.3 / 62.5 & 6.7 / 6.7\\
    \textbf{o4-mini}
      & \textbf{medium effort + tools} & 77.7 / 78.8 & 25.0 / 25.0\\
    \midrule
    \textbf{Claude Sonnet 4}
      & \textbf{medium} & 60.2 / 72.5 & 5.2 / 5.2\\
    \textbf{Claude Sonnet 4}
      & \textbf{medium + tools} & 55.0 / 59.2 & 6.9 / 6.9\\
    \midrule
    \textbf{Gemini 2.5 Pro}
      & \textbf{medium effort} & 66.0 / 66.0 & 4.2 / 4.2\\
    \textbf{Gemini 2.5 Pro}
      & \textbf{medium effort + tools} & 60.4 / 60.4 & 5.8 / 5.8\\
    \midrule
    \midrule
    \textbf{GPT-4o}
      & \textbf{no tools} & 14.6 / 14.6 & 0.0 / 0.0\\
    \textbf{GPT-4o}
      & \textbf{with tools} & 8.3 / 13.1 & 0.2 / 0.2\\
    \midrule
    \textbf{Llama 4 Scout}
      & \textbf{no tools} & 6.7 / 8.5 & 0.0 / 0.0\\
    \midrule
    \textbf{Qwen 2.5 VL 72B}
      & \textbf{no tools} & 9.2 / 10.0 & 0.0 / 0.0\\
    \bottomrule
  \end{tabularx}
\end{table}

\section{Rule-Grid Alignment} \label{app:RuleGridAlignment}
To investigate how accurately the natural-language rules reflect the models’ underlying reasoning, our team manually evaluated the solutions generated by o3, Claude Sonnet 4, and Gemini 2.5 Pro in the medium-effort + tools setting for two additional features: rule-grid alignment and visual errors. Rule-grid alignment describes cases where the model’s stated rule accurately captures the transformation demonstrated in its output grid. Visual errors refer to cases in visual settings where a grid’s incorrectness or inconsistency with its corresponding rule appears to stem, at least in part, from failures in the model’s perceptual apparatus rather than from a mismatch between its generated rule and underlying reasoning. Such errors commonly involved failures to recognize the exact grid dimensions, slight inaccuracies in object placement, or incorrect mappings between colors and their numerical encodings. 

For these evaluations, one team member provided an initial judgment for each task, after which any ambiguous cases were discussed as a group until consensus was reached. The results of these evaluations are presented in \autoref{tab:rgd_ve}. We find that the natural-language rules align with their corresponding grids in the vast majority of cases; across all evaluated models and settings, agreement exceeded 90\% of tasks. This supports the view that the proposed rules generally reflect the reasoning used to produce the grid solutions. 
In addition, the relatively high rate of visual errors corroborates our earlier observations of failure modes in visual settings. Even the best-performing model and setting exhibited a visual error rate approaching 50\%, despite maintaining a high degree of rule-grid alignment. This suggests that the low grid accuracies observed in visual settings may be driven in large part by perceptual limitations, rather than solely by differences in reasoning capacity between visual and textual modalities.

\begin{table}[t]
  \caption{Percentage of tasks in which models exhibited either ``rule--grid alignment'' (RGA)—the generated rule accurately described the generated output grid—or, in visual settings, a ``visual error'' (VE). The medium-effort + tools setting was used for all models. Tasks with missing or invalidly formatted output grids and/or non-responsive or unclear rules are excluded from these percentages.}
  \label{tab:rgd_ve}
  \centering
    \setlength{\tabcolsep}{4.5pt}
    \renewcommand{\arraystretch}{1.1}
    \begin{sc}
      \begin{tabular}{lccc}
        \toprule
      \textbf{Model} &
      \multicolumn{1}{c}{\textbf{Textual}} &
      \multicolumn{2}{c}{\textbf{Visual}} \\
      \textbf{(Medium + Tools)} & & & \\[-0.8em]
      \cmidrule(lr){2-2}\cmidrule(lr){3-4}
        & \textbf{RGA} & \textbf{RGA} & \textbf{VE} \\
        \midrule
        \textbf{o3}     & 98.1 & 97.4 & 49.1 \\
        \textbf{Claude} & 92.4 & 96.6 & 59.7 \\
        \textbf{Gemini} & 94.5 & 93.6 & 77.3 \\
        \bottomrule
      \end{tabular}
    \end{sc}
\vspace{-0.3cm}
\end{table}

\section{Error-Type Overview \label{app:error-type-overview}}
\begin{figure}[h]
  \centering
  \includegraphics[width=0.8\textwidth]{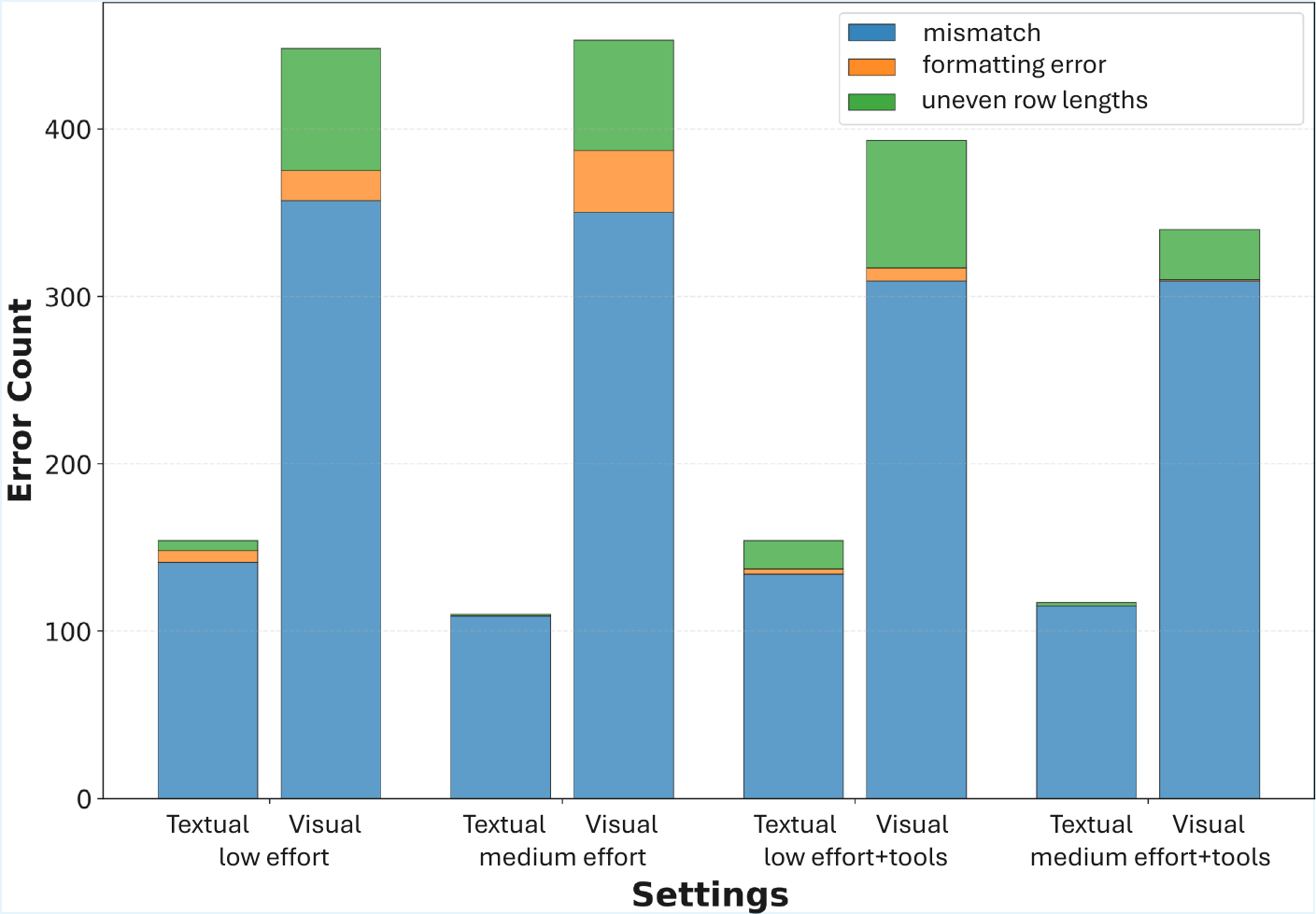}
  \caption{Overview of different error types for o3 in different experimental settings. For each setting, the left bar shows textual modality results and the right bar shows visual modality results. The most common error type is a simple mismatch error, in which the output grid and the ground truth grid are not identical, including cases of incorrect grid dimensions and single-cell mismatches. We also encountered parsing errors, most often originating from incorrect formatting (see \autoref{app:alternative_grid_formats})as well as from grids with rows of differing lengths.}
  \label{fig:Error-Type-Overview}
\end{figure}

\section{Examples of Correct-Unintended Rules \label{app:examples-correct-unintended-rules}}

This appendix provides concrete illustrations of some of the recurring shortcut types observed in model-generated rules. For each category, we give a brief description of the pattern and at least one representative example. These categories are neither mutually exclusive nor exhaustive; many rules combine elements of more than one shortcut type, and some shortcuts are too specific to a particular task or concept group to be usefully categorized. As such, this appendix gives a broad overview of just some of the most common shortcut modes observed.

\subsection{Exploiting Numerical Properties of the Integer Color Encoding
\label{app:shortcut-numerical}}

\begin{figure}[h]
    \centering
    \includegraphics[width=\textwidth]{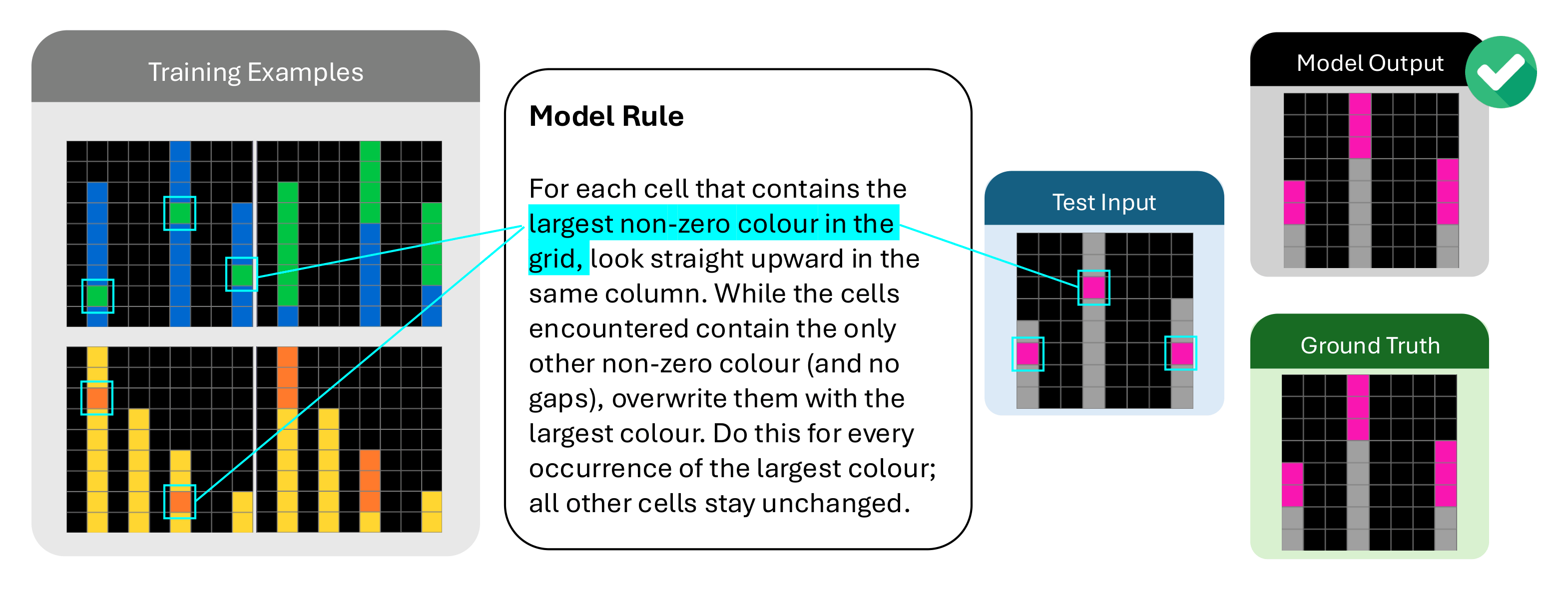}
    \caption{Correct-unintended rule exploiting numerical properties of the integer color encoding. o3, textual, low effort with tools, Above vs.\ Below (AboveBelow5, Test 1).}
    \label{fig:example-numerical-encoding}
\end{figure}

Models frequently exploit attributes of the integer color encoding, such as the ordering or even-odd parity of numbers, to find spurious correlations that explain the demonstrations. In the following example from the concept group Above vs.\ Below, shown in \autoref{fig:example-numerical-encoding}, o3 proposes the rule: "For each cell that contains the largest non-zero colour in the grid, look straight upward in the same column. While the cells encountered contain the only other non-zero colour (and no gaps), overwrite them with the largest colour. Do this for every occurrence of the largest colour; all other cells stay unchanged." Rather than identifying the secondary color as a spatial outlier, the model exploits the fact that, by coincidence, the outlier color has been assigned a higher integer value than the primary color in both training demonstrations. This pattern appears particularly common in tasks where colors must be selected or overwritten according to a spatial rule.

\subsection{Using Local Neighborhood Connectivity as a Proxy for 
Higher-Level Spatial Concepts
\label{app:shortcut-connectivity}}

Models sometimes use notions of local neighborhood connectivity to approximate higher-level spatial concepts, such as objectness or containment within another shape. Rather than identifying objects and their spatial relationships directly, such rules check for adjacency (or, in the case of containment, connectivity to the grid border) as a proxy for these concepts. 

\begin{figure}[h]
    \centering
    \includegraphics[width=\textwidth]{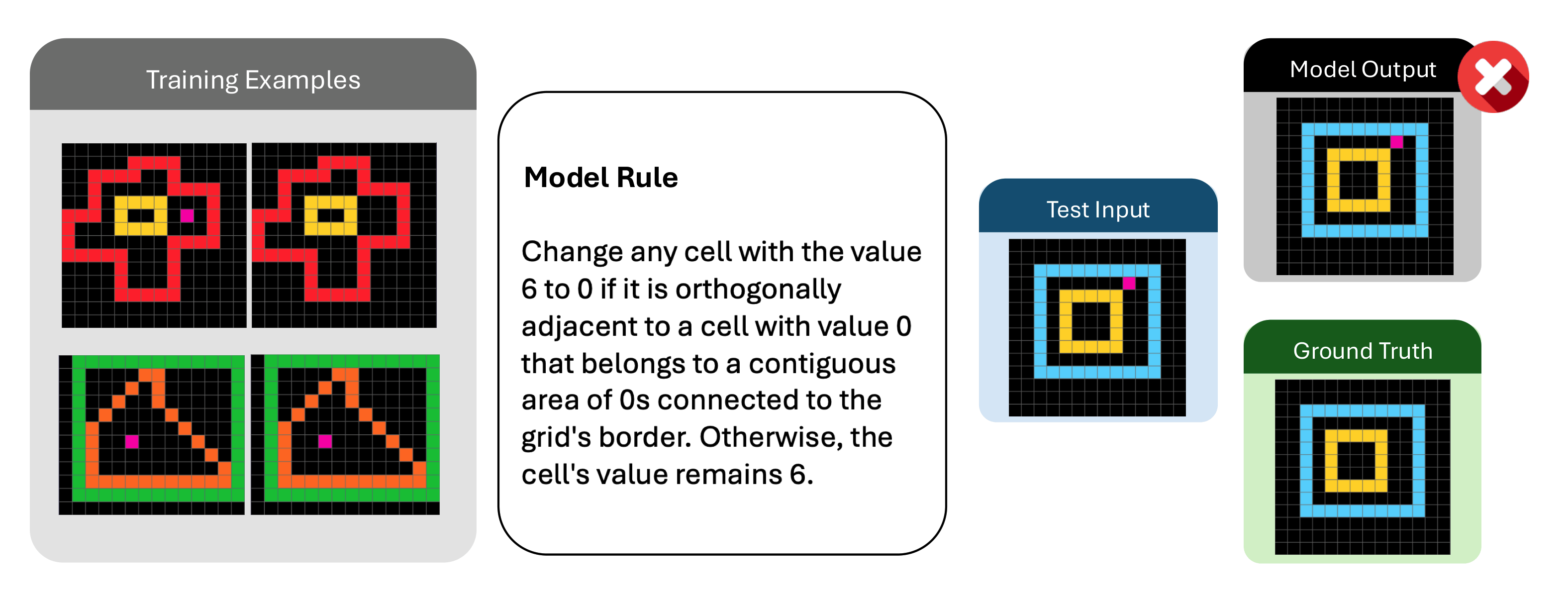}
    \caption{Rule using local neighborhood connectivity as a proxy for containment. Gemini 2.5 Pro, textual, medium effort, Inside vs.\ Outside (InsideOutside4, Test 1).}
    \label{fig:example-connectivity}
\end{figure}

In the example from Inside vs.\ Outside shown in \autoref{fig:example-connectivity}, Gemini 2.5 proposes the rule: "Change any cell with the value 6 [magenta] to 0 [black] if it is orthogonally adjacent to a cell with value 0 that belongs to a contiguous area of 0s connected to the grid's border. Otherwise, the cell's value remains 6." By checking for a potential path from adjacent cells to the grid edge, the model attempts to derive information about containment within another shape by instead using local connectivity.

\begin{figure}[h]
    \centering
    \includegraphics[width=\textwidth]{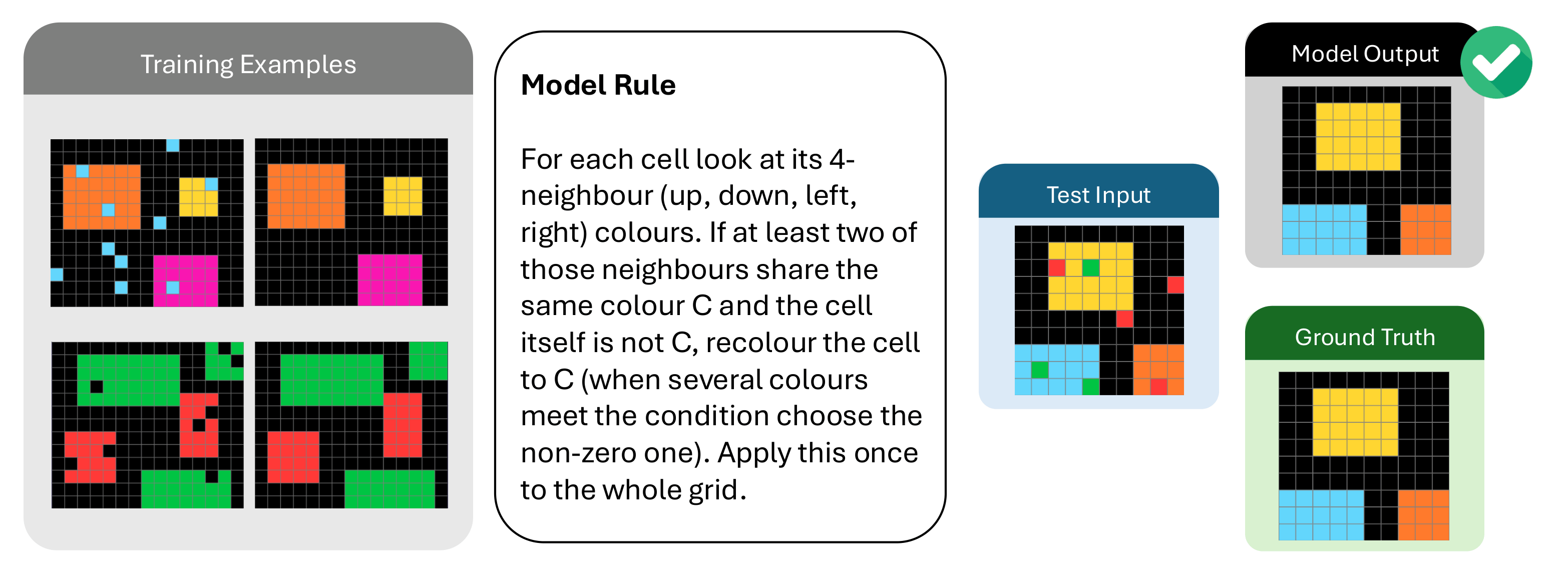}
    \caption{Correct-unintended rule using local neighborhood connectivity as a proxy for objectness. o3, textual, low effort with tools, CleanUp (CleanUp3, Test 2).}
    \label{fig:example-neighborhood}
\end{figure}

In the following example from the concept group CleanUp, shown in \autoref{fig:example-neighborhood}, o3 determines how to recolor each cell based on the majority color among its four orthogonal neighbors, rather than recognizing the larger coherent objects that should be preserved and removing the "noise" cells that surround them. This local majority-voting approach approximates the intended transformation in some cases but fails to capture the intended distinction between ``object'' and ``noise''.

\subsection{Over-Specified Algorithmic Rules
\label{app:shortcut-overspecified}}

\begin{figure}[h]
    \centering
    \includegraphics[width=\textwidth]{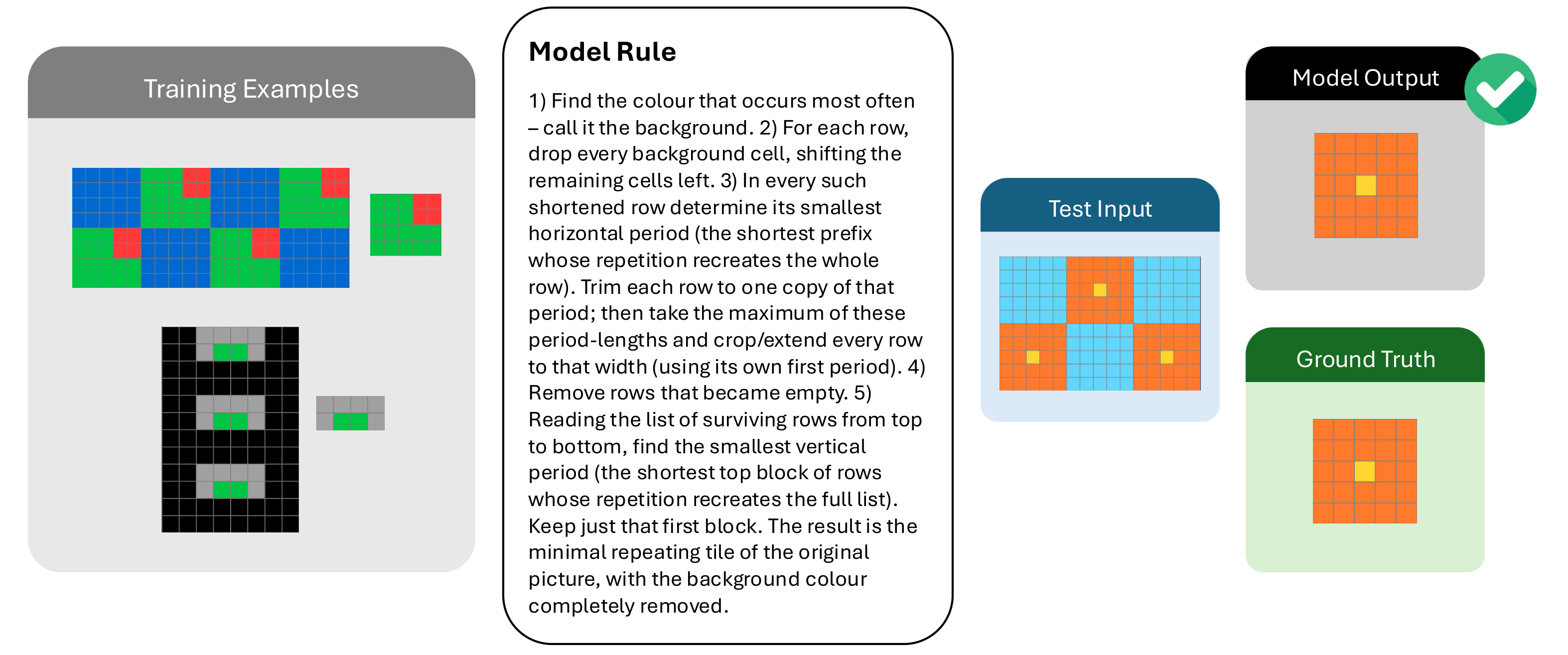}
    \caption{Correct-unintended algorithmic rule. o3, visual, medium effort with tools, ExtractObjects (ExtractObjects10, Test 1).}
    \label{fig:example-algorithmic-rule}
\end{figure}

Rather than describing a transformation in terms of the spatial or conceptual relationships it involves, models sometimes produce rules that are defined in mathematical or algorithmic ways that bear no resemblance to the intended abstraction. Such rules may produce the correct output on the given demonstrations while obscuring the underlying concept entirely. In the following example from the ExtractObjects group, shown in \autoref{fig:example-algorithmic-rule}, o3 proposes the rule: "1) Find the colour that occurs most often – call it the background. 2) For each row, drop every background cell, shifting the remaining cells left. 3) In every such shortened row determine its smallest horizontal period (the shortest prefix whose repetition recreates the whole row). Trim each row to one copy of that period; then take the maximum of these period-lengths and crop/extend every row to that width (using its own first period). 4) Remove rows that became empty. 5) Reading the list of surviving rows from top to bottom, find the smallest vertical period (the shortest top block of rows whose repetition recreates the full list). Keep just that first block." This rule is algorithmically well-defined, and indeed correctly isolates a single instance of the repeating pattern in each of the demonstrations, but bears no relation to the intended concept. Such rules appear to arise when simpler pattern-matching approaches have been exhausted, and the models resort to searching for more complex descriptions that fit the demonstrations.

\subsection{Proxy Heuristics Approximating the Intended Concept
\label{app:shortcut-proxy}}

\begin{figure}[h]
    \centering
    \includegraphics[width=\textwidth]{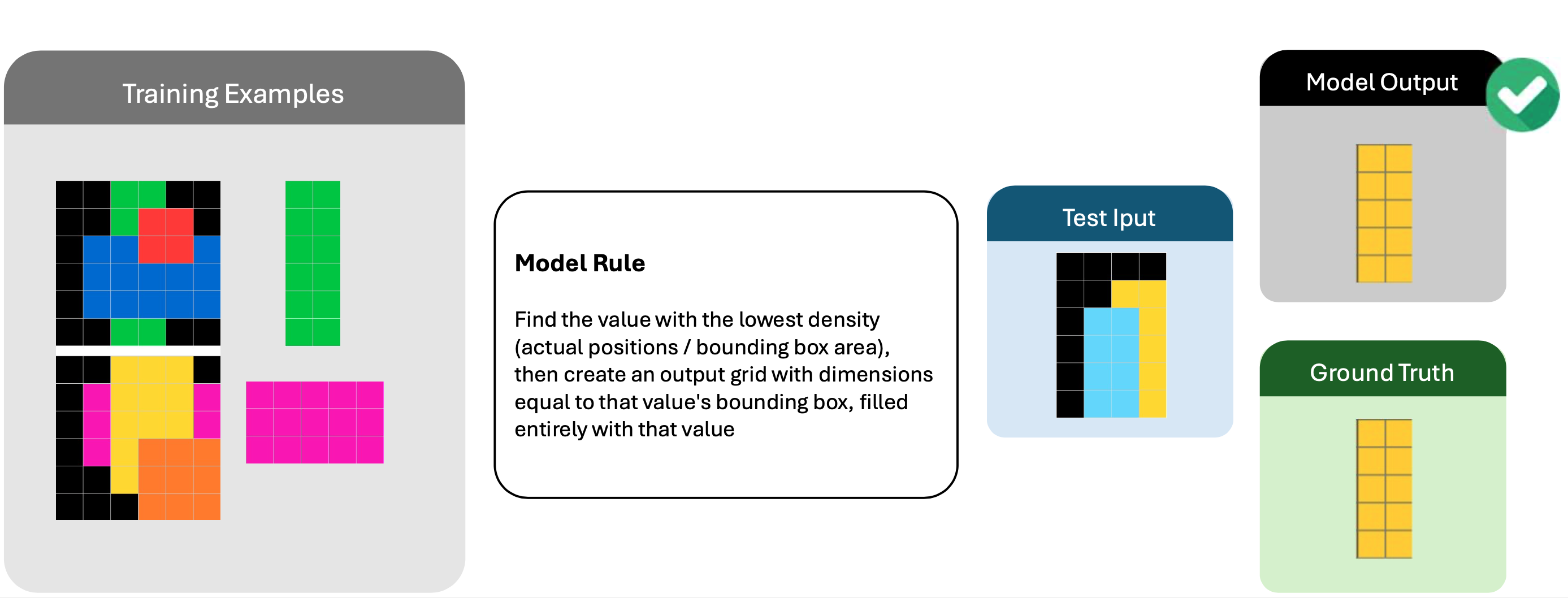}
    \caption{Correct-unintended rule using a density heuristic as a proxy for 3D spatial position. Claude Sonnet 4, textual, medium effort with tools, Top vs.\ Bottom 3D (TopBottom3D6, Test 1).}
    \label{fig:example-heuristic}
\end{figure}

Models sometimes identify a feature of the task that reliably correlates with the correct answer across the demonstrations without capturing the underlying concept. In the following example from the Top vs. Bottom 3D concept group, shown in \autoref{fig:example-heuristic}, Claude Sonnet 4 uses a density heuristic to approximate the most overlapped figure. The model conflates the bottommost shape in a 3D stack with the color that has the lowest density (number of cells divided by the area of its bounding box). This approach indeed works for the demonstrations, and would even work if applied to the other test examples for this task (that is, alternative test inputs that have the same demonstrations and underlying transformation rule). However, it fails to capture the notion of the bottommost shape in a 3-dimensional stack, and would not generalize to very similar variations where the correlation between density and position is violated.

\subsection{Case-by-Case or Demonstration-Specific Rules
\label{app:shortcut-casebycases}}

\begin{figure}[h]
    \centering
    \includegraphics[width=\textwidth]{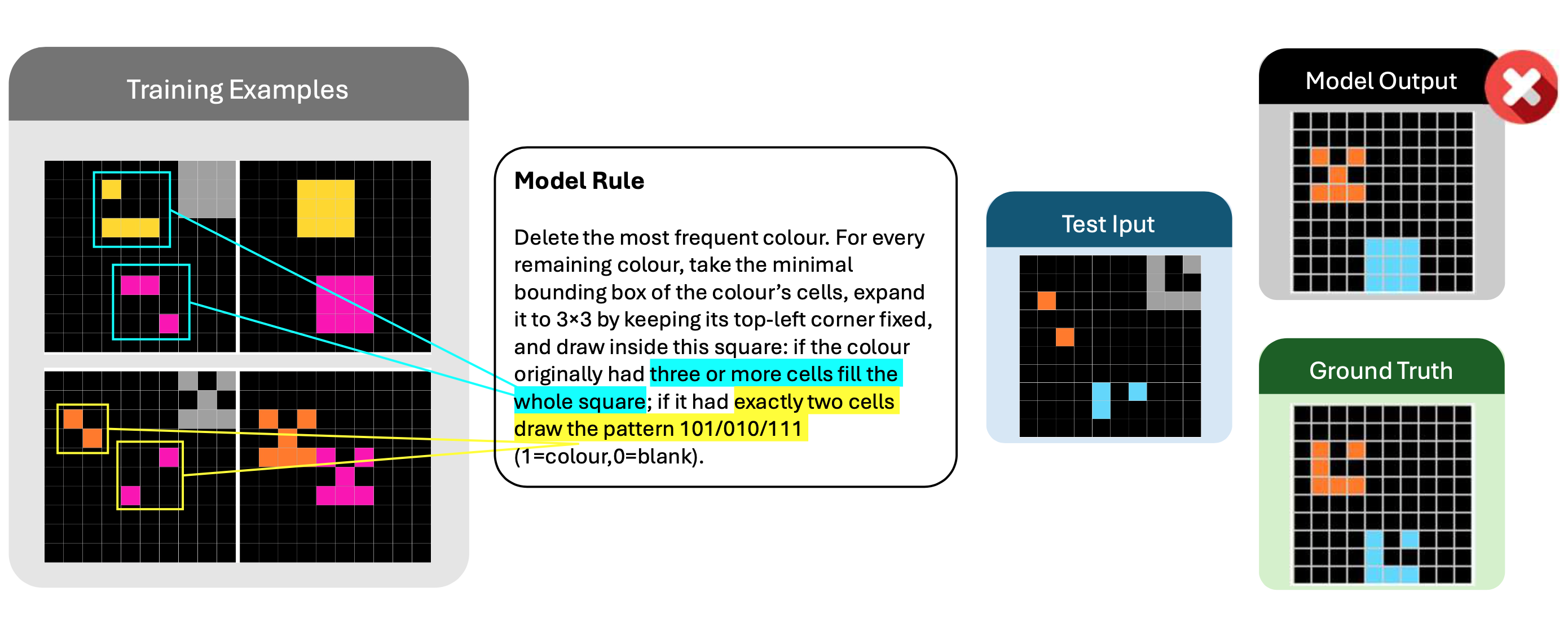}
    \caption{Correct-unintended, demonstration-specific rule. o3, medium effort with tools, Complete Shape (CompleteShape8, Test 1).}
    \label{fig:example-case-by-case}
\end{figure}

Rather than inferring a single rule that generalizes across demonstrations, models sometimes produce rules that amount to a disjoint set of cases, one per demonstration, instead of a single generalizing principle. While related to overfitting, the failure here is not necessarily about spurious features; the model may accurately describe each demonstration, while still failing to abstract a general rule from them. Such rules may nevertheless produce a correct output on the test case by coincidence. In the following example from the Complete Shape concept group, shown in \autoref{fig:example-case-by-case}, o3 does not recognize the relationship between the colored output shape and the gray prototype, and instead correlates the final shape with the number of colored cells in each demonstration, ignoring the role of the gray template completely.

This example also illustrates the proxy heuristics pattern described above in \autoref{app:shortcut-proxy}. The model's rule, to delete the most frequent color, coincidentally identifies the gray template in the given demonstrations, since gray happens to be the most frequent color in each. The co-occurrence of these two shortcut modes in a single rule helps illustrate why the categories in this appendix are not mutually exclusive.

\subsection{Exploiting Task Structure to Constrain the Solution Space
\label{app:shortcut-taskstructure}}

\begin{figure}[h]
    \centering
    \includegraphics[width=\textwidth]{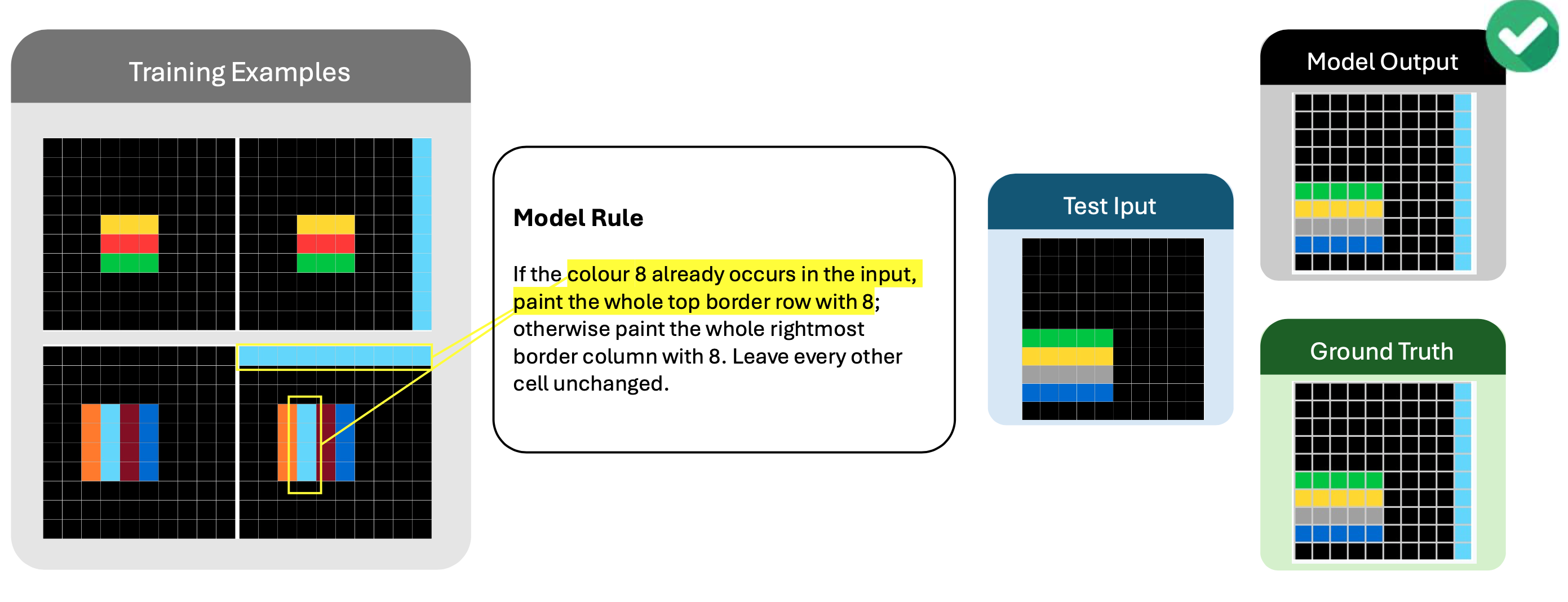}
    \caption{Correct-unintended rule exploiting constrained solution space. o3, textual, low effort, Horizontal vs.\ Vertical (HorizontalVertical10, Test 1.}
    \label{fig:example-constrained-solutions}
\end{figure}

For certain tasks, the structure of the demonstrations is such that only a small number of outputs are plausible, regardless of whether the underlying rule has been correctly identified. In such cases, models sometimes generate rules that exploit this constraint without identifying the concepts that actually determine the correct answer, producing correct outputs at rates that reflect the size of the plausible solution space rather than a true understanding of the task. In the following example from the Horizontal vs. Vertical concept group, shown in \autoref{fig:example-constrained-solutions}, o3 does not recognize the relation between the orientation of the colored shape components and the cyan row, and instead focuses on whether a cyan ("8") pixel already appears in the grid. Because the constrained solution space appears to limit plausible outputs to a cyan line at the top or on the right, a rule that identifies this constraint without identifying the principle that determines which of the two possibilities is correct will still produce the correct output roughly half the time. This makes it difficult to distinguish correct abstraction from lucky guessing based on the output grid alone, particularly when evaluating a small number of test cases.

In the visual modality, many of these same patterns are present, but can in some cases be difficult to disentangle from perceptual errors, since a misread grid can produce what appears to be a shortcut rule but is in fact a downstream consequence of incorrect perception. 

\newpage
\section{Effects of Varying Reasoning Effort and Tool Use on o3's Generated Rules}
\label{app:o3RuleAnalysis}

\begin{figure}[t]
  \centering
 \includegraphics[width=\textwidth]{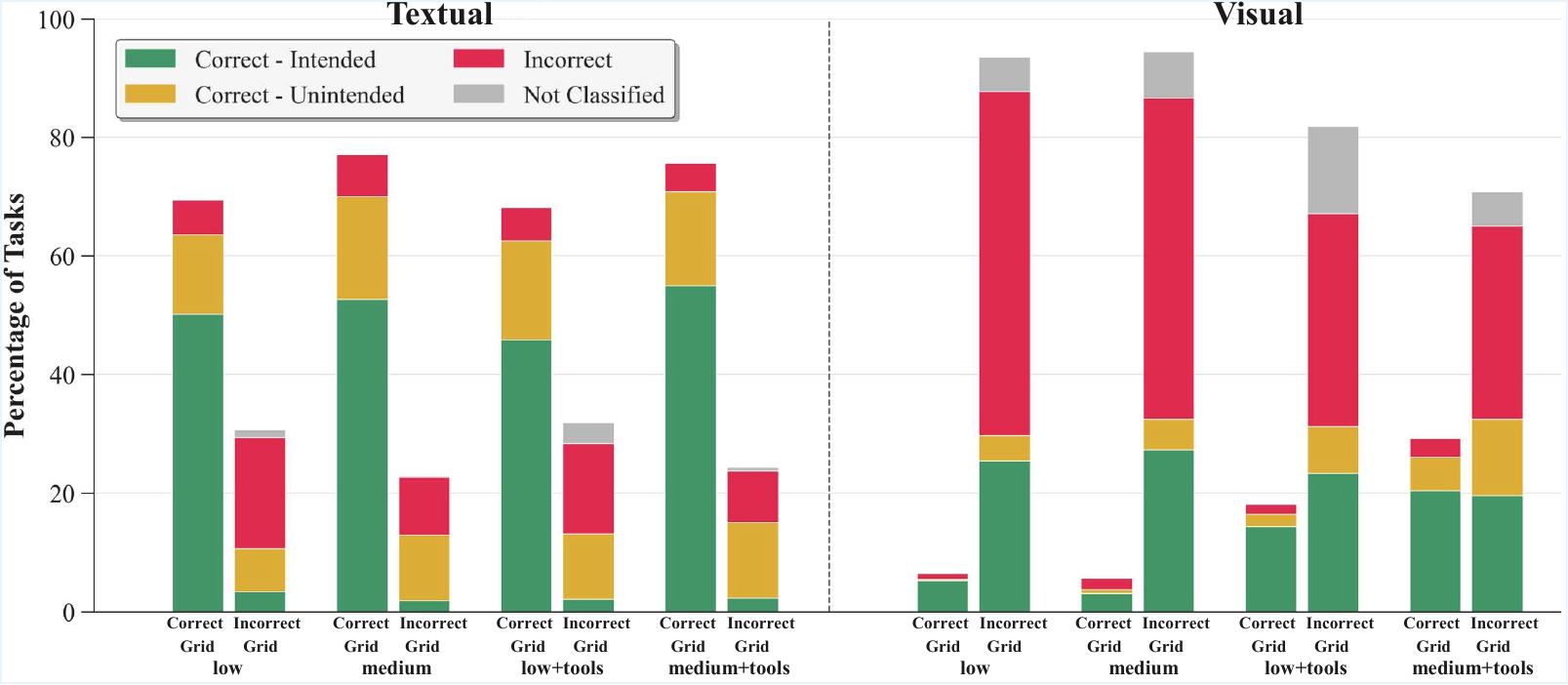}
  \caption{Results of rule evaluations for o3 across all settings. As in \autoref{fig:rule_eval_plot}, two bars showing the percentage of correct and incorrect output grids are included for each setting, with each bar showing the fraction of tasks for which the generated rule is correct-intended, correct-unintended, and incorrect. The actual percentages corresponding to regions on the bars are given in \autoref{app:rule_evaluation_data}. 
  \label{fig:rule_analysis_o3_settings}}
\end{figure}

\autoref{fig:rule_analysis_o3_settings} shows the effects of varying reasoning effort and tool use for the o3 model, in both textual and visual modalities. There are a few important observations to make.  First, in the visual setting, increasing reasoning effort from low to medium alone does not have any substantial effect on output accuracy or rule correctness, which aligns with prior work suggesting that test-time scaling does not have the dramatic effects in visual modalities that have been seen in text-only models \citep{hao2025can}.  However, enabling Python tool use does result in substantial improvement in output accuracy and rule correctness, especially at medium reasoning effort, likely because the model is able to use computer vision libraries. In contrast, in the textual setting, increasing reasoning effort has a larger positive effect on both output accuracy and rule correctness than enabling Python tool use. 
\section{Repeated-Run Analysis}
\label{app:repeated-runs}

To assess the variability of our pass@1 results, we evaluated o3, Claude Sonnet 4, and Gemini 2.5 Pro three times in the textual medium-effort setting, without tools. The subset consisted of all three test inputs for the first task in each of the 16 ConceptARC concept groups, yielding 48 test cases and 144 responses per model.

We calculated 95\% percentile-bootstrap confidence intervals using 100,000 resamples with replacement, resampling the observed responses within each test case. Output-grid accuracy included all responses, with missing and malformed outputs counted as incorrect. Rule-category rates exclude test cases with one or more nonresponses from that model.

\begin{table}[t]
  \centering
  \caption{Repeated-run results. All entries are percentages; bracketed values are 95\% percentile-bootstrap confidence intervals. Rule-category rates are computed excluding nonresponses.}
  \label{tab:repeated-run-results}
  \small
  \begin{tabular}{lccc}
    \toprule
    & \textbf{o3} & \textbf{Claude} & \textbf{Gemini} \\
    \midrule
    \multicolumn{4}{l}{\textit{Aggregate rates across three passes}} \\
    Output-grid accuracy & 74.3 [70.8, 77.8] & 61.1 [55.6, 66.7] & 68.8 [64.6, 72.9] \\
    Correct-intended rule & 59.9 [54.9, 64.8] & 59.0 [54.2, 63.9] & 56.1 [51.5, 60.6] \\
    Correct-unintended rule & 31.7 [26.8, 36.6] & 19.4 [15.3, 23.6] & 36.4 [31.8, 40.9] \\
    Incorrect rule & 8.5 [4.9, 12.0] & 18.8 [15.3, 22.9] & 7.6 [3.8, 11.4] \\
    Unclear rule & 0.0 [0.0, 0.0] & 2.8 [0.7, 4.9] & 0.0 [0.0, 0.0] \\
    \midrule
    \multicolumn{4}{l}{\textit{Item-level stability}} \\
    Same output correctness in all three passes & 81.3 & 52.1 & 72.9 \\
    Same rule label in all three passes & 54.3 & 54.2 & 56.4 \\
    Mean pairwise rule-label agreement & 68.1 & 66.0 & 70.9 \\
    \bottomrule
  \end{tabular}
\end{table}
\section{Data for Rule Evaluation Plots \label{app:rule_evaluation_data}}

The tables below give the data used to create \autoref{fig:rule_eval_plot} and \autoref{fig:rule_analysis_o3_settings}. For o3, Claude, Gemini, and human-generated rules, each cell in each table reports the percentage of rules assigned to each classification (Correct-Intended, Correct-Unintended, Incorrect, or Not-Classified) in each bar shown in \autoref{fig:rule_eval_plot} and \autoref{fig:rule_analysis_o3_settings}. Percentages are computed separately for cases with correct output grids and for cases with incorrect output grids. For each model, the results are reported for both modalities. The Not-Classified percentages include missing rules as well as rules that were unresponsive or unclear. Rules were not collected for incorrect grids in the original human experiment, and so all human responses with incorrect grids are reported here as Not-Classified; these statistics are estimates based on the 73\% grid accuracy reported by the original experimenters. The final row of statistics for humans show the rule classification breakdown when excluding Not-Classified rules.

\begin{table}[h]
  \caption{Data used to create \autoref{fig:rule_eval_plot}}
  \label{tab:rule_bar_fractions}
  \centering
  \small
  \setlength{\tabcolsep}{2pt}
  \renewcommand{\arraystretch}{1.25}
  \begin{tabularx}{\textwidth}{lYYYYYYYY}
    \toprule
    \textbf{Model (Modality)} &
    \multicolumn{4}{c}{\textbf{Correct Grid}} &
    \multicolumn{4}{c}{\textbf{Incorrect Grid}} \\
    \cmidrule(lr){2-5}\cmidrule(lr){6-9}
    & \thead{Correct-\\Intended} & \thead{Correct-\\Unint.} & \thead{Incorrect} & \thead{Not- \\Classif.}
    & \thead{Correct-\\Intended} & \thead{Correct-\\Unint.} & \thead{Incorrect}  & \thead{Not- \\Classif.} \\
    \midrule
    \textbf{o3 (Textual)}
     & 72.7 & 20.9 & 6.3 & 0.0 & 9.4 & 52.1 & 35.9 & 2.6 \\
    \textbf{o3 (Visual)}
     & 70.0 & 19.3 & 10.7 & 0.0 & 27.6 & 18.2 & 45.9 & 8.2 \\
    \midrule
    \textbf{Claude Sonnet 4 (Textual)}
     & 80.3 & 9.5 & 9.1 & 1.1 & 29.6 & 20.8 & 36.1 & 13.4 \\
    \textbf{Claude Sonnet 4 (Visual)}
     & 57.6 & 6.1 & 33.3 & 3.0 & 11.2 & 2.7 & 71.4 & 14.8 \\
    \midrule
    \textbf{Gemini 2.5 Pro (Textual)}
     & 72.4 & 19.7 & 7.9 & 0.0 & 5.8 & 25.3 & 65.3 & 3.7 \\
    \textbf{Gemini 2.5 Pro (Visual)}
     & 78.6 & 3.6 & 17.9 & 0.0 & 20.1 & 5.8 & 71.2 & 2.9 \\
     \midrule
    \textbf{Human (All Data)}
     & 74.0 & 3.2 & 4.1 & 18.8 & -- & -- & -- & 100.0 \\
    \textbf{Human (Excl. Not-Classified)}
     & 91.1 & 3.9 & 5.0 & -- & -- & -- & -- & -- \\
    \bottomrule
  \end{tabularx}
\end{table}

\begin{table}[htbp]
  \caption{Data used to create \autoref{fig:rule_analysis_o3_settings}.}
  \label{tab:o3_setting_bar_fractions}
  \centering
  \small
  \setlength{\tabcolsep}{2pt}
  \renewcommand{\arraystretch}{1.25}
  \begin{tabularx}{\textwidth}{lYYYYYYYY}
    \toprule
    \textbf{o3 Setting (Modality)} &
    \multicolumn{4}{c}{\textbf{Correct Grid}} &
    \multicolumn{4}{c}{\textbf{Incorrect Grid}} \\
    \cmidrule(lr){2-5}\cmidrule(lr){6-9}
    & \thead{Correct-\\Intended} & \thead{Correct-\\Unint.} & \thead{Incorrect} & \thead{Not- \\Classif.}
    & \thead{Correct-\\Intended} & \thead{Correct-\\Unint.} & \thead{Incorrect}  & \thead{Not- \\Classif.} \\
    \midrule
    \textbf{Low effort (Textual)}
      & 72.3 & 19.2 & 8.5 & 0.0 & 13.2 & 23.7 & 59.2 & 3.9 \\
    \textbf{Low effort (Visual)}
      & 81.3 & 3.1 & 15.6 & 0.0 & 28.1 & 4.7 & 60.7 & 6.5  \\
    \midrule
    \textbf{Medium effort (Textual)}
      & 68.4 & 22.4 & 9.2 & 0.0 & 8.2 & 48.2 & 42.7 & 0.9 \\
    \textbf{Medium effort (Visual)}
      & 55.6 & 11.1 & 33.3 & 0.0 & 28.9 & 5.5 & 57.4 & 8.2 \\
    \midrule
    \textbf{Low effort + tools (Textual)}
      & 67.5 & 24.2 & 8.3 & 0.0 & 6.5 & 35.1 & 47.4 & 11.0 \\
    \textbf{Low effort + tools (Visual)}
      & 79.3 & 11.5 & 9.2 & 0.0 & 28.5 & 9.7 & 43.8 & 18.1 \\
    \midrule
    \textbf{Medium effort + tools (Textual)}
      & 72.7 & 20.9 & 6.3 & 0.0 & 9.4 & 52.1 & 35.9 & 2.6 \\
    \textbf{Medium effort + tools (Visual)}
      & 70.0 & 19.3 & 10.7 & 0.0 & 27.6 & 18.2 & 45.9 & 8.2 \\
    \bottomrule
  \end{tabularx}
\end{table}

\newpage

\section{Concept Performance Overview \label{app:concept_performance}}

ConceptARC \citep{moskvichev2023conceptarc} is organized around 16 basic spatial and semantic concepts.  Each concept group consists of 10 tasks that focus on the concept in different ways, with each task containing three distinct test grids.  \autoref{tab:Concept_Textual} and \autoref{tab:Concept_Visual} give the per-concept-group accuracies (each out of 30 grids) of the reasoning models we evaluated (using medium reasoning effort and Python tools), as well as human accuracies on these concept groups from \citet{moskvichev2023conceptarc}.  Human trials were administered solely using visual images of the demonstration and test grids, but human accuracy is reported in both tables for ease of comparison.  

\begin{table}[htbp]
  \caption{\textbf{Concept performance (Textual):} Per-concept accuracy (\%) on Concept-ARC using medium effort + tools. Highest accuracy per concept is shown in bold.}
  \label{tab:Concept_Textual}
  \centering
  \small
  \setlength{\tabcolsep}{6pt}
  \renewcommand{\arraystretch}{1.1}
  \begin{tabularx}{\textwidth}{lYYYYY}
    \toprule
    \textbf{Concept} & \textbf{Gemini 2.5 Pro} & \textbf{o3} & \textbf{o4-mini} & \textbf{Claude Sonnet 4} & \textbf{Human} \\
    \midrule
    AboveBelow & 60 & \textbf{90} & 83.3 & 63.3 & 69 \\
    Center & 70 & 93.3 & \textbf{96.7} & 83.3 & 84 \\
    CleanUp & 23.3 & 46.7 & 60 & 46.7 & \textbf{89} \\
    CompleteShape & 56.7 & \textbf{70} & 66.7 & 50 & 71 \\
    Copy & 66.7 & 70 & \textbf{90} & 56.7 & 78 \\
    Count & \textbf{86.7} & 80 & 80 & 76.7 & 61 \\
    ExtendToBoundary & 60 & \textbf{90} & 83.3 & 50 & 81 \\
    ExtractObjects & 56.7 & 76.7 & \textbf{86.7} & 43.3 & 67 \\
    FilledNotFilled & 73.3 & 76.7 & \textbf{83.3} & 63.3 & 82 \\
    HorizontalVertical & 53.3 & \textbf{70} & \textbf{70} & 63.3 & 68 \\
    InsideOutside & 66.7 & \textbf{80} & 73.3 & 43.3 & 68 \\
    MoveToBoundary & 63.3 & \textbf{80} & 70 & 40 & 78 \\
    Order & 50 & 70 & 70 & 40 & \textbf{76} \\
    SameDifferent & 56.7 & 83.3 & \textbf{86.7} & 53.3 & 68 \\
    TopBottom2D & 76.7 & 86.7 & \textbf{93.3} & 56.7 & 79 \\
    TopBottom3D & 46.7 & 53.3 & 56.7 & 50 & \textbf{70} \\
    \bottomrule
  \end{tabularx}
\end{table}

\begin{table}[h]
  \caption{\textbf{Concept performance (Visual):} Per-concept accuracy (\%) on Concept-ARC using medium effort + tools. Highest accuracy per concept is shown in bold.}
  \label{tab:Concept_Visual}
  \centering
  \small
  \setlength{\tabcolsep}{6pt}
  \renewcommand{\arraystretch}{1.1}
  \begin{tabularx}{\textwidth}{lYYYYY}
    \toprule
    \textbf{Concept} & \textbf{Gemini 2.5 Pro} & \textbf{o3} & \textbf{o4-mini} & \textbf{Claude Sonnet 4} & \textbf{Human} \\
    \midrule
    AboveBelow & 0 & 20 & 10 & 0 & \textbf{69} \\
    Center & 6.7 & 43.3 & 26.7 & 6.7 & \textbf{84} \\
    CleanUp & 10 & 23.3 & 26.7 & 13.3 & \textbf{89} \\
    CompleteShape & 3.3 & 30 & 23.3 & 16.7 & \textbf{71} \\
    Copy & 3.3 & 20 & 23.3 & 3.3 & \textbf{78} \\
    Count & 16.7 & 53.3 & 50 & 0 & \textbf{61} \\
    ExtendToBoundary & 0 & 20 & 13.3 & 3.3 & \textbf{81} \\
    ExtractObjects & 3.3 & 30 & 36.7 & 0 & \textbf{67} \\
    FilledNotFilled & 6.7 & 30 & 20 & 0 & \textbf{82} \\
    HorizontalVertical & 3.3 & 33.3 & 20 & 6.7 & \textbf{68} \\
    InsideOutside & 6.7 & 16.7 & 13.3 & 10 & \textbf{68} \\
    MoveToBoundary & 3.3 & 30 & 10 & 16.7 & \textbf{76} \\
    Order & 10 & 33.3 & 36.7 & 13.3 & \textbf{60} \\
    SameDifferent & 6.7 & 26.7 & 26.7 & 6.7 & \textbf{68} \\
    TopBottom2D & 13.3 & 33.3 & 50 & 3.3 & \textbf{79} \\
    TopBottom3D & 0 & 23.3 & 13.3 & 10 & \textbf{70} \\
    \bottomrule
  \end{tabularx}
\end{table}

\newpage

\begin{figure}[htbp]
  \centering
 \includegraphics[width=\textwidth]{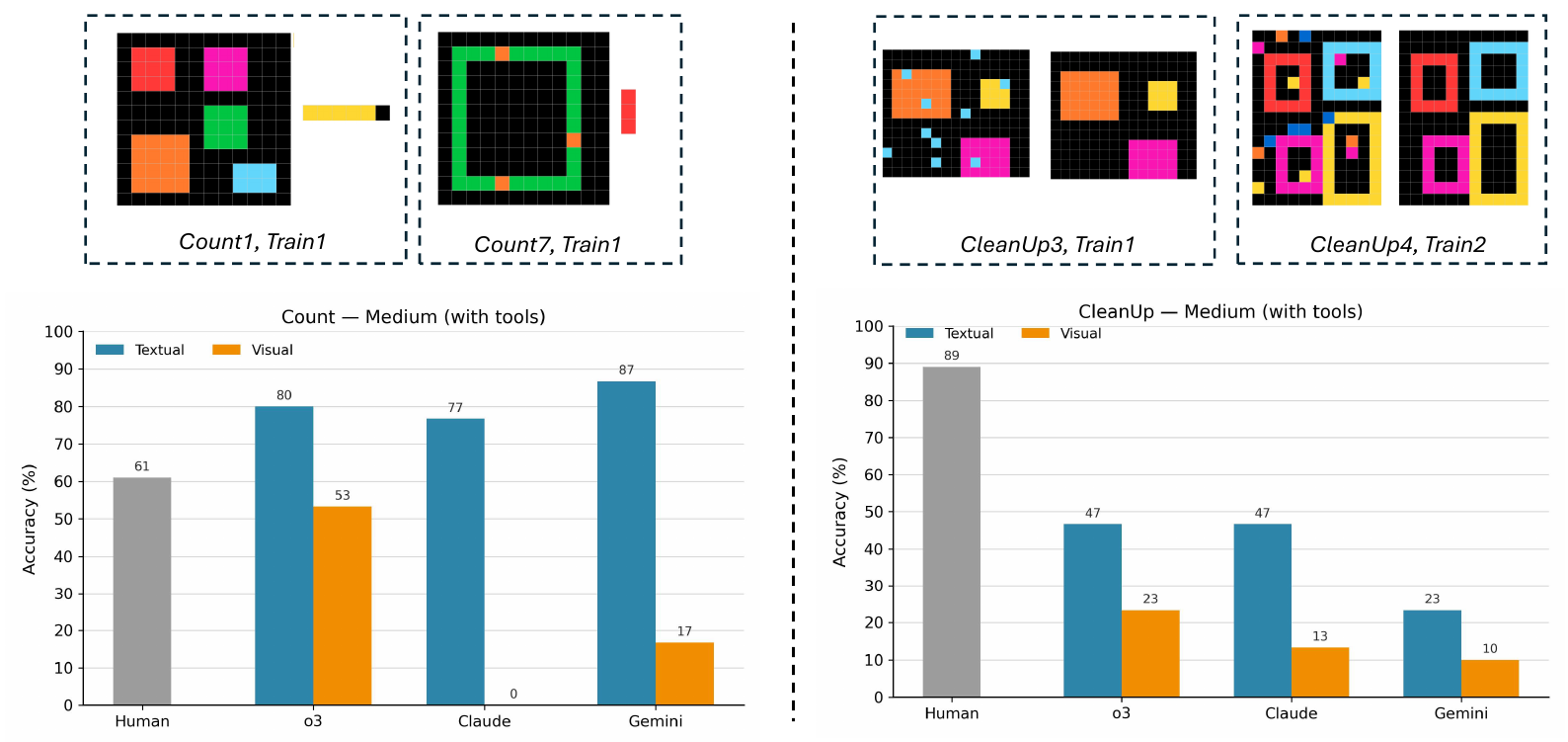}
  \caption{Top row: Two example demonstrations from the concepts with the highest and lowest gaps between human and model performance, Count and CleanUp. Bottom row: Concept-wise output grid accuracy across three reasoning models in a medium with tools setting (note that we compare against the strongest settings in \autoref{app:Concept_diff}).
  \label{fig:concept_overview_plot}}
\end{figure}

\subsection{Concept Difficulty Evaluation} \label{app:Concept_diff}
Although we find no significant correlation in concept difficulty across modalities (visual vs. textual) or with human participants, we do identify several overarching trends. Full per-concept performance comparisons are shown in \autoref{tab:Concept_Textual} and \autoref{tab:Concept_Visual}; in particular, note the performance differences for the concepts ``Count'' and ``CleanUp.''Tasks in the group ``Count'' frequently involve the production of simple, singular output rows or columns, denoting the count of specific characteristics (e.g shapes, colors, corners). Correspondingly, output grids are often small and easy to generate. In the visual modality, this is the performance closest to humans for both o3 (-7.7\%)\footnote{Note that for this and the following comparisons we consider the strongest model settings, which is not necessarily medium with tools} and Gemini (-44\%), and in the textual modality, this concept also results in the biggest positive difference (o3: +32.3\%; Gemini: +25.7\%; Claude: +15.7\%). In contrast, tasks in the CleanUp concept group require the removal of several colors, shapes, or isolated pixels, as well as full reproduction of the remaining input grid. In this concept group, even o3 using medium effort + tools is significantly outperformed by human participants in the visual setting (-65.7\%). Similarly, answers to CleanUp tasks constitute the largest negative performance gap in the textual modality (-46.3\%). The gap between the other models is even larger. This is a strong indicator that, regardless of modality, models struggle significantly with producing complex output grids.

\newpage
\section{Comparison of Concepts Used in Human and AI Model Rules: Details and Additional Results} 
\label{app:ConceptUseComparison}

To create \autoref{fig:per_llm_proportions_with_tools} in the main text, we first preprocessed the human- and AI-model-generated rules by (1) converting all text to lower case, (2) removing hyphens, apostrophes, and other punctuation,  and  (3) lemmatizing words (e.g., ``filling'', ``filled'', ``fills'' $\to$ ``fill'') using the Python package \texttt{spaCy}. We also standardized spelling variants using a custom mapping (e.g., ``neighbour $\to$ neighbor'', ``colour $\to$ color'', and ``centre $\to$ center''). After preprocessing, we searched all human- and AI-model-generated rules to determine the proportion containing at least one relevant term. 

We also extended the results shown in \autoref{fig:per_llm_proportions_with_tools} in the main paper in two ways: (1) Comparing use of objectness-related terms and grid-specific terms in humans and AI models across our rule-classification categories (Correct-Intended, Correct-Unintended, and Incorrect), and (2) 
performing a similar comparison for each of ConceptARC's 16 concept groups. 

\begin{figure}[ht]
  \centering
\includegraphics[width=\columnwidth]{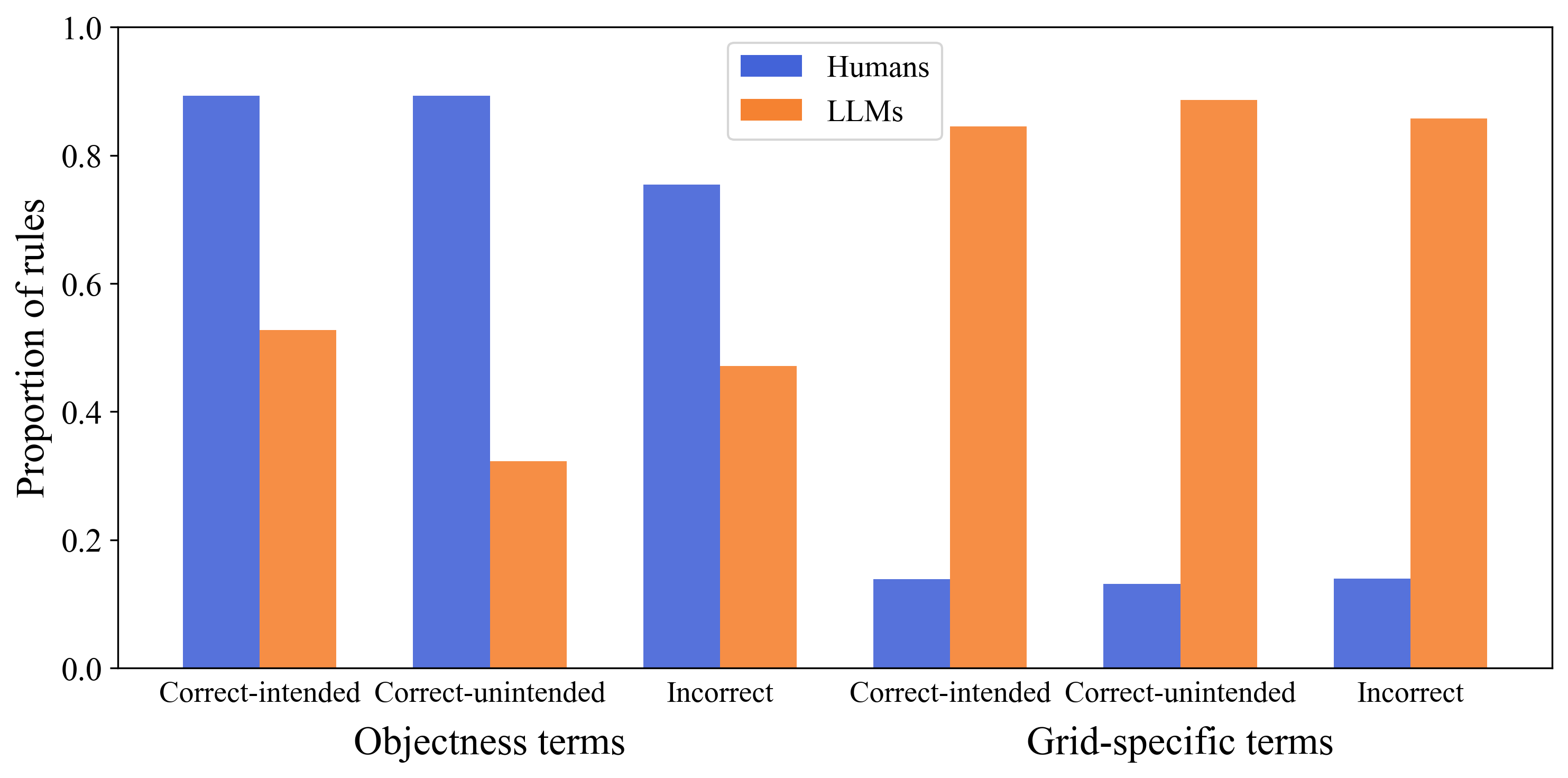}
 \caption{For each classification (correct-intended, correct-unintended, incorrect), proportion of human- and LLM-generated rules that used objectness-related and grid-specific terms. For humans, there were 2212 correct-intended rules, 122 correct-unintended rules, and 122 incorrect rules in total. For AI models, there were 686 correct-intended rules, 158 correct-unintended rules, and 70 incorrect rules in total.}
  \label{fig:per_correctness_proportions_with_tools}
\end{figure} 

For (1), we used the same terms as in Figure~5 in the main text, but separate out proportion of rules in which these terms were used in each type of rule.  The results are shown in \autoref{fig:per_correctness_proportions_with_tools}. The figure shows that, even across these different categories, humans are substantially more likely to use objectness-related terms than AI models. It also shows a notable drop in the use of objectness-related terms in the correct-unintended rules generated by models, but not in those generated by humans, suggesting that the ways humans deviate from intended solutions may still differ substantially from the kinds of unintended shortcuts discovered by models.

For (2), for each concept group, we collected terms from both human and AI model rules that we judged to express the relevant abstract concept, and then applied the same  preprocessing and analysis described in the main paper. In addition, since ``copy'' is one of the terms being used, we excluded irrelevant occurrences of ``copy" that often appeared in the first sentence of human-generated rules (e.g., ``Press the Copy button''), which referred to the interface used by the human participants rather than to the problem being solved.    

\begin{figure}[ht]
  \centering
\includegraphics[width=1\columnwidth]{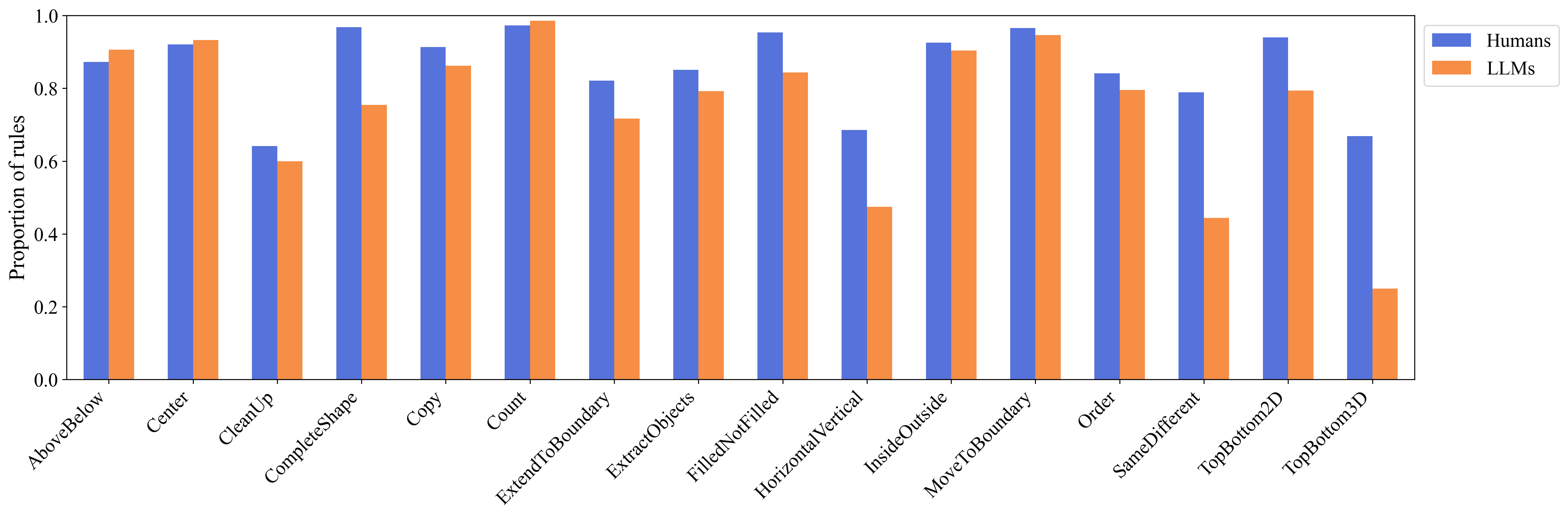}
  \caption{For each concept group, proportion of rules containing at least one concept-associated term, for humans and for AI models (o3, Claude, and Gemini combined).  The concept-associated terms  are: \\  
  \textbf{AboveBelow:} above, below, over, under, top, bottom, up, down, upward, downward, beneath, underneath, lower, upper, higher, topmost, bottommost, uppermost, high, low. \\
\textbf{Center:} center, middle, midpoint, median, central, between, equidistant, halfway, centermost, midway. \\
\textbf{CleanUp:} clean, remove, complete, noise, noisy, random, randomly, scatter, disparate, outlier, extra, anomalous, nonconforming, deviation, isolate, interrupt, mismatch, clutter, delete, replace, eliminate, declutter, miss, break, unbroken, solid. \\
\textbf{CompleteShape:} complete, shape, square, miss, entire, gap, fill, finish, repair, replace, fix, break, incomplete, unbroken, full, pattern, hole, connect, unfinished. \\
\textbf{Copy:} copy, second, double, recreate, duplicate, match, repeat, same, identical, replicate. \\
\textbf{Count:} count, amount, total, for each, for every, how many, however many, number of, that many. \\
\textbf{ExtendToBoundary:} extend, boundary, reach, touch, edge, hit, meet, toward, towards, top, bottom, end, side, far, furthest, wall, stop, stretch, contact, border, barrier, extension, topmost, bottommost, leftmost, rightmost. \\
\textbf{ExtractObjects:} extract, object, shape, rectangle, rectangular, copy, recreate, replicate, reproduce, duplicate, crop. \\
\textbf{FilledNotFilled:} fill, outline, inside, middle, interior, border, enclose, surround, hollow, contain, hole, solid, solidly, inner, interior, exterior, empty, miss, gap, open, frame, internal, external, space, within, outer, full. \\
\textbf{HorizontalVertical:} horizontal, vertical, orient, orientation, left to right, top to bottom, up and down. \\
\textbf{InsideOutside:} inside, outside, in, out, within, border, frame, enclose, contain, inner, outer, interior, exterior, innermost, outermost, internal, external. \\
\textbf{MoveToBoundary:} move, boundary, touch, slide, toward, border, contact, edge, side, wall, reach, shift, bottom, top, barrier, topmost, bottommost, hit, meet, top, bottom, end, far, furthest, stop, contact, border, leftmost, rightmost, adjacent. \\
\textbf{Order:} order, small, large, bottom, top, ascend, count, sort, arrange, rearrange, reorder, reorganize, sequence, big, descend, higher, lower, high, low, increase, decrease, least, great, few, most. \\
\textbf{SameDifferent:} same, different, copy, match, identical, unique, repeat, differ, duplicate, unique, replicate. \\
\textbf{TopBottom2D:} top, bottom, on, topmost, bottommost, under, over, lower, low, higher, high, upper, uppermost, lowermost, above, below, underneath, beneath. \\
\textbf{TopBottom3D:} top, bottom, behind, front, back, forward, cover, under, over, underneath, above, below, topmost, bottommost, uppermost, lowermost, beneath, overlay, foreground, background. \\
} \label{fig:per_concept_proportions_with_tools}
\end{figure}

\autoref{fig:per_concept_proportions_with_tools} shows the results comparing humans (2456 rules) with the three AI models (914 rules from the three models combined), using the concept-related terms specified in the caption.  The plot shows that humans tended to use more concept-specific terms than LLMs in 13 of the 16 concepts, although in some cases the difference is small.  As noted in the main paper, these results remain preliminary and reflect subjective choices in the selection of key terms; developing more principled and automated ways to identify concept-related terms is a direction for future work.   

\newpage

\section{Correct-Intended Coverage}
\begin{table}[htbp]
  \caption{\textbf{Correct-intended task coverage:} Number of tasks covered correctly by category and modality, with coverage rates listed as a percentage of the 480 total ConceptARC tasks. Here, a task is considered ``covered'' if the model in question produced a correct-intended rule in \textit{any} of its solutions for that task in the given modality. The ``AnyModel'' rows show task coverage aggregated across all three reasoning models, and the entry for humans shows the coverage of tasks for which at least one human subject produced a correct-intended rule.}
  \label{tab:CorrectIntendedCoverage}
  \centering
  \small
  \setlength{\tabcolsep}{6pt}
  \renewcommand{\arraystretch}{1.1}
  \begin{tabularx}{\textwidth}{l l Y Y}
    \toprule
    \textbf{Category} & \textbf{Modality} & \textbf{Covered} & \textbf{Percentage} \\
    \midrule
    Humans & Overall & 476 & 99.17 \\
    o3 & Textual & 412 & 85.83 \\
    o3 & Visual & 281 & 58.54 \\
    Claude & Textual & 346 & 72.08 \\
    Claude & Visual & 83 & 17.29 \\
    Gemini & Textual & 326 & 67.92 \\
    Gemini & Visual & 140 & 29.27 \\
    AnyModel & Textual & 454 & 94.58 \\
    AnyModel & Visual & 320 & 66.67 \\
    \bottomrule
  \end{tabularx}
\end{table}

\subsection{Correct-Intended Coverage Implications}
\autoref{tab:CorrectIntendedCoverage} shows that, while models (in textual modality) all have a decent coverage, pooling their answers only leads to a moderate increase as compared to the best performing single model (+9\%). While the overall coverage is notably lower in visual modality, the increase when pooling the three models again is comparable to textual (+8\%). As we do not have individual human performance data, we unfortunately cannot compute similar statistics for pooling single human performances. However, these results again point out stronger abstractive reasoning abilities in a human panel, which only failed to derive the correct abstract transformation in 4 test examples (AboveBelow2 Test 3, CompleteShape4 Test 1, HorizontalVertical5 Test 1, Order9 Test 3). It is worth noting that our human rule dataset did not contain rules for grids with incorrect outputs, meaning that the score of 476/480 is a lower bound, as some humans who produced incorrect grids may nevertheless have inferred the correct-intended rules.
\section{Distribution of Correct-Unintended Rules Across Concepts}
\label{app:ConceptDistribution}
Upon discovering the usage of correct but unintended rules, we were interested in analyzing the distribution of these among different concepts. In particular, models might be systematically employing unintended rules on tasks they lack meaningful priors for. Under textual modalities, when models arrived at a ``correct'' rule, they produced correct-unintended results in about 29.82\% those cases with a standard deviation of 16.44\%. The concepts with the highest share of correct-unintended rules were \textit{TopBottom3D} (70.59\%), \textit{CleanUp} (52.27\%) and \textit{HorizontalVertical} (45.12\%). For the visual modality, the average usage of correct-unintended rules was 27.0\%, with a standard deviation of 13.92\%. Again, the concepts with the highest share are \textit{TopBottom3D} (62.50\%) and \textit{CleanUp} (40\%), but also \textit{SameDifferent} (47.83\%). \textit{HorizontalVertical} had a reduced share of unintended rules, with only 35.42\%, ranking at fourth-most. \newline

We generally refer to the share of unintended abstractions of correct rules here (intended and unintended), rather than of all rules, in order to account for the difference in rule correctness between modalities. As \autoref{tab:Concept_Textual} and \autoref{tab:Concept_Visual} show, \textit{TopBottom3D} is one of the most difficult concepts for models when measured by accuracy (second-lowest in textual, third-lowest in visual), so it is not surprising that models largely rely on unintended rules when solving corresponding tasks. Analyzing the rules generated for tasks in this concept group more closely, few of them actually addressed 3-dimensional arrangement, but instead relied on shallower features, such as density or bounding-box interceptions. 

While \textit{CleanUp} produced the lowest overall accuracy in textual modality, it achieved the fourth-highest in visual modality, so it is intuitive to find increased shortcut-usage using text representations. The high share of unintended rules in vision-based inputs is somewhat surprising, but is likely due to difficulties with recognizing global patterns. This is an issue with both modalities, as models tend to employ local patterns, or case-by-case logic, which were overfit to the training demonstrations. In particular, models frequently recognize simple line-based patterns, such as alternating horizontal or vertical lines, but struggle with recognizing individual objects,since they seem to lack a strong prior for ``objectness.''

In the other concept groups, and on various tasks in general, we found that the models' generated rules involve recurring heuristics, including the employment of bounding boxes, four/eight-neighbor connectivity, as well as finding paths to the grid edge or object boundary by stepping through adjacent (4- or 8-connected) cells. These unintended abstractions seemed to be part of a general-purpose tool-box, which models employed for various purposes and not specific to single concepts.
\section{Output Accuracy Without Task ``Order5''} \label{app:WithoutOrder5}

We discovered that in the ConceptARC corpus \citep{moskvichev2023conceptarc}, one task (task 5 of the``Order'' concept group) includes a training demonstration containing a misplaced grid cell. We reanalyzed the data in \autoref{tab:arc_results_modality} with Order5 removed and obtained the results shown in \autoref{tab:order5_removed}. The resulting changes are negligible and do not meaningfully affect our conclusions.

\begin{table}[h]
  \caption{This table reports the same data as \autoref{tab:arc_results_modality}, with Order5 removed from the analysis. \label{tab:order5_removed}}
  
  \centering
  \begin{small}
    \setlength{\tabcolsep}{4.5pt}
    \renewcommand{\arraystretch}{1.1}
    \begin{sc}
      \begin{tabular}{l@{\hspace{6pt}}c@{\hspace{8pt}}cccc}
        \toprule
        \textbf{Model} &  &
        \textbf{low} &
        \textbf{medium} &
        \textbf{low +T} &
        \textbf{medium +T} \\
        \midrule
        \textbf{o3} & \textnormal{t} &
        68.6 & 77.6 & 68.3 & 75.9 \\
                    & \textnormal{v} &
        6.7  & 5.7           & 18.2 & 29.4 \\
        \midrule
        \textbf{Claude} & \textnormal{t} &
        N/A & 60.6 & N/A & 55.3 \\
                        & \textnormal{v} &
        N/A & 5.2  & N/A & 6.9 \\
        \midrule
        \textbf{Gemini} & \textnormal{t} &
        N/A & 66.5 & N/A & 60.4 \\
                        & \textnormal{v} &
        N/A & 4.2  & N/A & 5.9 \\
        \bottomrule
      \end{tabular}
    \end{sc}
  \end{small}
  \vskip -0.1in
\end{table}

\section{Compute Requirements \label{app:compute}}

All experiments using proprietary frontier models were conducted through commercial cloud-based APIs (OpenAI for o3, o4-mini, and GPT-4o; Anthropic for Claude Sonnet 4; and Google for Gemini 2.5 Pro). As such, the underlying compute, memory, and storage requirements are not disclosed to users. We therefore report the reproducible compute information available to us: the API provider, model, number of calls,  reasoning/effort setting, and other user-set hyperparameters.

Each model-setting evaluation consisted of 480 API calls, one for each ConceptARC task. The experimental results reported in \autoref{fig:rule_eval_plot} and \autoref{fig:rule_analysis_o3_settings} comprise 12 model-setting evaluations, for a total of $480 \times 12 = 5760$ API calls. Of these, 8 evaluations used a medium-effort reasoning setting (6 of which had tools enabled) and 4 used a low-effort reasoning setting (2 of which had tools enabled). 4 additional model-setting evaluations were completed for reasoning models (Claude Sonnet 4 and Gemini 2.5 Pro), all with medium-effort reasoning and without tools enabled, but were not rule-annotated in this study; these contribute an additional $480 \times 4 = 1920$ API calls. As noted in \autoref{sec:Methodology}, Gemini 2.5 Pro and Claude Sonnet 4 were evaluated with a reasoning budget of 16K tokens, chosen to approximate OpenAI’s medium-effort API setting.

Apart from minor API debugging and prompt-validation calls, no preliminary or failed experiments were conducted for this submission.
\newpage
\section{Preliminary Experiments to Assess Overfitting on ARC-AGI-1 and ARC-AGI-2 \label{app:ARC_AGI_Overfitting}}

As we described in the main paper, the  textual inputs we employed for ConceptARC tasks was the same numerical encoding that is used in all ARC-Prize evaluations and in most other published evaluations on ARC-AGI-1 \cite{ARC2025Github} and ARC-AGI-2 \cite{ARC2Github}

After noticing many examples in which AI models used unintended features based on this numerical encoding, we decided to do a simple test to see if AI models have  overfitted to this encoding.  For each task in the public evaluation sets for ARC-AGI-1 and ARC-AGI-2, we simply substituted symbols for numbers in the arrays representing grids:
\begin{quotation}{}
\noindent 0 $\rightarrow$ ! \\
1 $\rightarrow$ ? \\
2 $\rightarrow$ \& \\
3 $\rightarrow$ @ \\
4 $\rightarrow$ \# \\
5 $\rightarrow$ \% \\
6 $\rightarrow$ \$ \\
7 $\rightarrow$ \textasciicircum \\
8 $\rightarrow$ \textasciitilde \\
9 $\rightarrow$ \slash \\
\end{quotation}

In a preliminary study, we tested OpenAI's GPT-5.2 model (at medium effort, temperature 1.0) using the original numerical encoding and the new symbol encoding, and with the same prompt used in \cite{chollet2024o3}. The accuracies for two independent runs on each evaluation set are given in \autoref{tab:arc_agi_overfitting}.  As the table shows, for both datasets on each pass, the accuracy for the symbol encoding is substantially lower than that for the original numerical encoding. Since the textual encoding is meant to be arbitrary, these results suggest that, due to training on ARC tasks, models such as GPT-5.2 have overfitted to the numerical encoding (and may be using numerical ``shortcuts'' similar to those we described in the main paper) and thus their zero-shot performance using this encoding overestimates their more general zero-shot performance.  

\begin{table}[t]
\centering
\caption{Accuracies of GPT-5.2 (with medium reasoning effort) on numeric and symbol encodings for ARC-AGI public evaluation sets.  For each setting, accuracies of two independent passes are given. \\ \label{tab:arc_agi_overfitting}}
\begin{tabular}{|c|c|c|} \hline
Dataset & Numerical & Symbol \\ \hline
ARC-AGI-1 (pass 1) & 77.2\% & 63.2\% \\ \hline
ARC-AGI-1 (pass 2) & 80.8\% & 67.1\% \\ \hline
ARC-AGI-2 (pass 1) & 22.5\% & 16.81\% \\ \hline
ARC-AGI-2 (pass 2) & 23.47\% & 13.47\% \\ \hline
\end{tabular}
\end{table}

\newpage
\section*{NeurIPS Paper Checklist}

\begin{enumerate}

\item {\bf Claims}
    \item[] Question: Do the main claims made in the abstract and introduction accurately reflect the paper's contributions and scope?
    \item[] Answer: \answerYes{}
    \item[] Justification: \justification{The claims in the abstract and introduction explicitly spell out the paper's contributions and scope.}

\item {\bf Limitations}
    \item[] Question: Does the paper discuss the limitations of the work performed by the authors?
    \item[] Answer: \answerYes{} 
    \item[] Justification: \justification{Section 8 describes the limitations.}

\item {\bf Theory assumptions and proofs}
    \item[] Question: For each theoretical result, does the paper provide the full set of assumptions and a complete (and correct) proof?
    \item[] Answer: \answerNA{} 
    \item[] Justification: \justification{The paper does not include theoretical results.}

    \item {\bf Experimental result reproducibility}
    \item[] Question: Does the paper fully disclose all the information needed to reproduce the main experimental results of the paper to the extent that it affects the main claims and/or conclusions of the paper (regardless of whether the code and data are provided or not)?
    \item[] Answer: \answerYes{} 
    \item[] Justification: \justification{The paper details the models, prompts, and hyperparameters used in the experiments, along with other details of the experimental methodology. For the human study, we use data from another published study, provided to us by one of the authors of that study. In Section 7 we give the URL for a repository with all of the data we collected and analyzed in the paper, including our manual rule evaluations, as well as the URL for a web interface for viewing the data and rule evaluations.}

\item {\bf Open access to data and code}
    \item[] Question: Does the paper provide open access to the data and code, with sufficient instructions to faithfully reproduce the main experimental results, as described in supplemental material?
    \item[] Answer: \answerYes{} 
    \item[] Justification: \justification{In Section 7 we give the URL for a repository with all of the data we collected and analyzed in the paper, as well as the URL for a web interface for viewing the data.}

\item {\bf Experimental setting/details}
    \item[] Question: Does the paper specify all the training and test details (e.g., data splits, hyperparameters, how they were chosen, type of optimizer) necessary to understand the results?
    \item[] Answer: \answerYes{} 
    \item[] Justification: \justification{In this paper we use commercial and open-weight pre-trained models, that are accessed through APIs.  We do not do any training or fine-tuning ourselves.  The models, prompts, and relevant hyperparameter settings are detailed in the paper.}

\item {\bf Experiment statistical significance}
    \item[] Question: Does the paper report error bars suitably and correctly defined or other appropriate information about the statistical significance of the experiments?
    \item[] Answer: \answerYes{} 
    \item[] Justification: \justification{The paper reports 95\% bootstrap confidence intervals for the repeated-run analysis, as well as paired confidence intervals and McNemar tests for the visual performance--competence gaps.}

\item {\bf Experiments compute resources}
    \item[] Question: For each experiment, does the paper provide sufficient information on the computer resources (type of compute workers, memory, time of execution) needed to reproduce the experiments?
    \item[] Answer: \answerYes{} 
    \item[] Justification: \justification{Appendix~\ref{app:compute} reports the API providers and models used, explains why underlying hardware details are unavailable for the hosted proprietary models, and provides the number of API calls and hyperparameters used for the experiments used in this paper.}
    
\item {\bf Code of ethics}
    \item[] Question: Does the research conducted in the paper conform, in every respect, with the NeurIPS Code of Ethics \url{https://neurips.cc/public/EthicsGuidelines}?
    \item[] Answer: \answerYes{} 
    \item[] Justification: \justification{We have read the NeurIPS Code of Ethics and believe our work is in compliance with it.  The human study data we analyze was shared with us from another study, which was judged ``exempt'' by a university IRB and which (in 2022) paid participants approximately \$12--\$15 per hour. All data from that study was anonymized; we did not receive any identifying information about the participants.}

\item {\bf Broader impacts}
    \item[] Question: Does the paper discuss both potential positive societal impacts and negative societal impacts of the work performed?
    \item[] Answer: \answerYes{} 
    \item[] Justification: \justification{In the Conclusions section we include a discussion of broader impacts.}
    
\item {\bf Safeguards}
    \item[] Question: Does the paper describe safeguards that have been put in place for responsible release of data or models that have a high risk for misuse (e.g., pre-trained language models, image generators, or scraped datasets)?
    \item[] Answer: \answerNA{} 
    \item[] Justification: \justification{Our paper does not include any release of new models, and to the best of our knowledge our data does not present any risk of misuse.}

\item {\bf Licenses for existing assets}
    \item[] Question: Are the creators or original owners of assets (e.g., code, data, models), used in the paper, properly credited and are the license and terms of use explicitly mentioned and properly respected?
    \item[] Answer: \answerYes{} 
    \item[] Justification: \justification{The authors of this paper are the creators/owners of all code and data used in this study, except for the human data that we obtained from the author of another ConceptARC study (credit for which is cited in \autoref{sec:Methodology}).  In the paper (and appendices) we provided information for all of the proprietary models (and open-weight models) we evaluated.}

\item {\bf New assets}
    \item[] Question: Are new assets introduced in the paper well documented and is the documentation provided alongside the assets?
    \item[] Answer: \answerYes{}{} 
    \item[] Justification: \justification{All of the code and data we generated in this study is publicly available at \url{https://huggingface.co/datasets/AIHumanAbstraction/ConceptARC_Rule_Annotations}; a web interface for viewing the data is publicly available at \url{https://aihumanabstraction.github.io/}.}

\item {\bf Crowdsourcing and research with human subjects}
    \item[] Question: For crowdsourcing experiments and research with human subjects, does the paper include the full text of instructions given to participants and screenshots, if applicable, as well as details about compensation (if any)? 
    \item[] Answer: \answerNo{} 
    \item[] Justification: \justification{We obtained human data from another study.  We cite the paper describing the details of that study, which includes the full text of instructions given to participants as well as details of compensation.}

\item {\bf Institutional review board (IRB) approvals or equivalent for research with human subjects}
    \item[] Question: Does the paper describe potential risks incurred by study participants, whether such risks were disclosed to the subjects, and whether Institutional Review Board (IRB) approvals (or an equivalent approval/review based on the requirements of your country or institution) were obtained?
    \item[] Answer: \answerNo{} 
    \item[] Justification: \justification{The paper we cite (from whose authors we obtained the human study data we analyzed) describes that the study was ruled ``exempt'' by a university IRB.}

\item {\bf Declaration of LLM usage}
    \item[] Question: Does the paper describe the usage of LLMs if it is an important, original, or non-standard component of the core methods in this research? Note that if the LLM is used only for writing, editing, or formatting purposes and does \emph{not} impact the core methodology, scientific rigor, or originality of the research, declaration is not required.
    \item[] Answer: \answerNA{} 
    \item[] Justification: \justification{The paper describes in detail our evaluation of LLMs on the ConceptARC benchmark, but we did not use LLMs to assist in the research reported here or in the writing of this paper.}

\end{enumerate}

\end{document}